\documentclass[11pt]{article}

\usepackage[final]{acl}
\usepackage[T1]{fontenc}           
\usepackage{times}                 
\usepackage{latexsym}              
\usepackage{inconsolata}           
\usepackage{textcomp}              
\usepackage{amssymb}
\usepackage{amsmath}               
\usepackage[table]{xcolor}         
\definecolor{tablegroupcolor}{gray}{0.9} 
\usepackage{booktabs}              
\usepackage{tabularx}              
\usepackage{makecell}              
\usepackage{multirow}              
\usepackage{siunitx}               
\usepackage{tabularx}
\usepackage{xltabular}            
\usepackage{graphicx}              
\usepackage{stfloats}              
\usepackage{enumitem}              
\usepackage[most]{tcolorbox}       
\usepackage{cuted} 

\newtcblisting[auto counter]{promptbox}[2][]{
    breakable,                      
    listing only,                   
    colback=white,                  
    colframe=black!60,              
    colbacktitle=black!60,          
    coltitle=white,                 
    fonttitle=\bfseries\rmfamily\large, 
    title={Box \thetcbcounter: #2},                     
    arc=2pt,                        
    boxrule=0.8pt,                  
    left=10pt, right=10pt, top=8pt, bottom=8pt,
    boxsep=0pt,
    toptitle=4pt, bottomtitle=4pt,
    listing options={
        basicstyle=\small\ttfamily,  
        breaklines=true,             
        breakindent=0pt,
        columns=fullflexible,        
        keepspaces=true,             
        lineskip=3pt,                
    },
    #1
}

\usepackage{microtype}             
\usepackage{hyperref}
\usepackage{url}
\usepackage[misc]{ifsym}
\usepackage{footmisc}
\DefineFNsymbols*{newfootnote}{%
    \textasteriskcentered{\textrm{\Letter}}\dag\textdaggerdbl{\ding{73}}\P{**}%
    {\ding{172}}{\ding{173}}{\ding{174}}}
\setfnsymbol{newfootnote}

\title{\textsc{\benchmark}: Benchmarking Long-Horizon Agentic Planning with Verifiable Constraints}

\newcommand{\qwenlogo}{\raisebox{0pt}{~\includegraphics[scale=0.01]{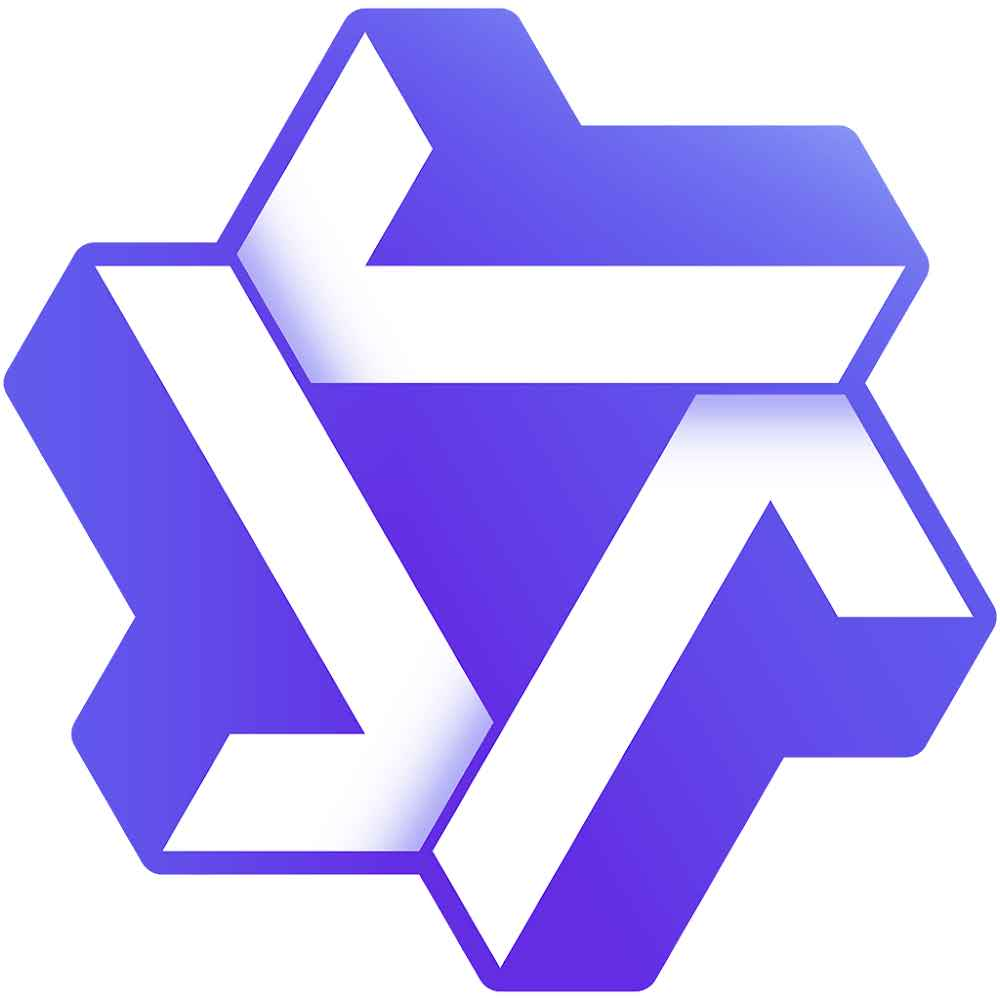}}~}
\author{
  Yinger Zhang\thanks{Equal contribution.},\quad
  Shutong Jiang\footnotemark[1],\quad
  Renhao Li\footnotemark[1],\quad
  Jianhong Tu\thanks{Corresponding author.}, \\
  \textbf{Yang Su,}\quad
  \textbf{Lianghao Deng,}\quad
  \textbf{Xudong Guo,}\quad
  \textbf{Chenxu Lv,}\quad
  \textbf{Junyang Lin\footnotemark[2]} \\[0.5em]
  Qwen Team, Alibaba Group\qwenlogo\\
  \texttt{\{tujianhong.tjh, junyang.ljy\}@alibaba-inc.com}\\[0.5em]
  \normalsize Homepage: \url{https://qwenlm.github.io/Qwen-Agent/en/benchmarks/deepplanning/}
}

\newcommand{\benchmark}{DeepPlanning}

\begin{document}
\maketitle
\begin{abstract}
While agent evaluation has shifted toward long-horizon tasks, most benchmarks still emphasize local, step-level reasoning rather than the global constrained optimization (e.g., time and financial budgets) that demands genuine planning ability. Meanwhile, existing LLM planning benchmarks underrepresent the active information gathering and fine-grained local constraints typical of real-world settings. To address this, we introduce \textsc{\benchmark}, a challenging benchmark for practical long-horizon agent planning. It features multi-day travel planning and multi-product shopping tasks that require proactive information acquisition, local constrained reasoning, and global constrained optimization. Evaluations on {\benchmark} show that even frontier agentic LLMs struggle with these problems, highlighting the importance of reliable explicit reasoning patterns and parallel tool use for achieving better effectiveness-efficiency trade-offs. Error analysis further points to promising directions for improving agentic LLMs over long planning horizons. We open-source the code and data to support future research.
\end{abstract}

\section{Introduction}
Agentic tool use has emerged as a fundamental capability for large language models (LLMs), extending their utility far beyond parametric knowledge. Recently, evaluation of agents has shifted from short-horizon, tool-centric benchmarks~\citep{li-etal-2023-api,qin2024toolllm,patil2025the} toward long-horizon, user-centric tasks~\citep{qian2025userbench,andrews2025scaling,he2025vitabench}. In these evolving settings, agents are expected to utilize tools not merely to execute isolated commands, but to satisfy complex, implicit user requests while adhering to strictly defined domain policies~\citep{yao2024tau,barres2025tau}.

Despite this progress, a critical gap remains in how these capabilities are evaluated. Existing tool-use benchmarks predominantly emphasize local, step-level constrained reasoning within individual actions, such as filtering hotels based on amenities or modifying a specific flight segment. While necessary, this perspective is insufficient for real-world complexity. Practical scenarios often impose global constrained optimization requirements, where constraints such as total time budgets, cumulative financial costs, and cross-subtask dependencies restrict the entire solution space rather than any single step. Current agent evaluation benchmarks largely fail to assess their ability to navigate these holistic boundaries, leaving the community without a reliable measure for comprehensive agent planning.

While efforts exist to benchmark LLMs on classical~\citep{valmeekam2023planbench,stechly2025on} and temporal planning tasks~\citep{zhang-etal-2024-timearena,ding-etal-2025-tcp}, these settings are simplified and abstract away the complex information acquisition process inherent to reality. In contrast, real-life long-horizon tasks such as multi-day travel and multi-product shopping require agents to proactively seek information from environments. However, existing benchmarks impede thorough evaluation: they suffer from either ineffective global constraints~\citep{yao2022webshop,lyu2025deepshop}, trivial local constraints~\citep{shao2025chinatravel}, or overly coarse global constraints~\citep{xie2024travelplanner,singh-etal-2024-personal}.

\begin{figure*}[!t]
  \centering
  \includegraphics[width=1.0\textwidth]{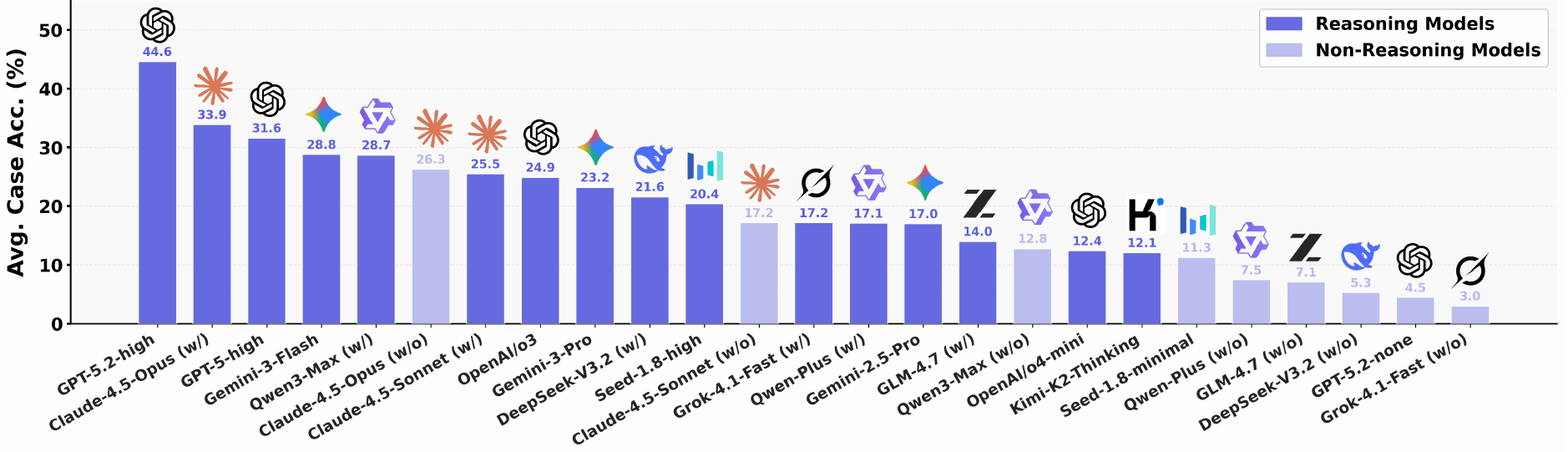} 
  \caption{Performance of frontier models on {\benchmark}, ranked by average case accuracy across \textit{Travel Planning} and \textit{Shopping Planning}. Dark/light bars denote reasoning versus non-reasoning models.}
  \label{fig:model_rank}
\end{figure*}

To address these limitations, we introduce \textsc{\benchmark}, a comprehensive benchmark grounded in challenging real-world tasks designed to evaluate practical LLM agents. We posit that a capable agent must integrate three key competencies: (i) \textbf{Proactive Information Acquisition}---the ability to actively search for and retrieve necessary state information from the environment; (ii) \textbf{Local Constrained Reasoning}---the ability to handle explicit and implicit logic within sub-tasks; and (iii) \textbf{Global Constrained Optimization}---the ability to optimize the overall solution under holistic constraints. Guided by this perspective, we build our benchmark across two complex real-world domains, \textit{Travel Planning} and \textit{Shopping Planning}, by constructing realistic yet challenging planning tasks through a comprehensive pipeline encompassing database and toolbox design, layered task generation, and manual quality control. 
We then evaluate frontier LLMs from multiple model families on {\benchmark} and expose clear limitations in their ability to solve these tasks (see Figure~\ref{fig:model_rank}). Our experiments show that reliable explicit reasoning patterns and parallel tool use are critical for achieving better effectiveness-efficiency trade-offs in complex planning. Detailed error analyses further underscore the importance of the three competencies: (i) even top-performing agents may omit necessary tool calls over long planning horizons; (ii) implicit environmental constraints (e.g., limited seat availability on flights or mismatched attraction opening hours) are harder to detect than explicit user requirements; and (iii) agents still lack robust global consistency checking and backtracking for long-horizon, tightly coupled tasks.

Our contributions are summarized as follows:
\begin{itemize}[leftmargin=*]
    \item We introduce {\benchmark}, a benchmark designed to challenge the long-horizon planning abilities of LLM agents through complex travel and shopping tasks. It supports reproducible, easy-to-verify evaluation via offline sandboxes and rule-based checkers.
    \item Our large-scale evaluation exposes fundamental limitations of frontier agentic LLMs in long-horizon planning, highlighting the need for more complex and realistic evaluation settings.
    \item Through comprehensive analysis, we identify key factors that improve effectiveness--efficiency trade-offs on {\benchmark} and outline promising directions for strengthening agentic LLMs over extended planning horizons.
\end{itemize}

\section{Related Works}
\subsection{Evaluation of Agentic Tool Use}
The inherent complexity of real-world scenarios is pushing LLMs beyond explicit, short-term tool use \citep{li-etal-2023-api,qin2024toolllm} toward more implicit, long-horizon tasks \citep{yao2024tau,barres2025tau,andrews2025scaling,qian2025userbench,he2025vitabench}. Another line of research evaluates agents' ability to solve complex objectives in noisy, web-based environments \citep{yao2022webshop,deng2023mindweb,zhou2024webarena,he-etal-2024-webvoyager,lyu2025deepshop,wei2025browsecomp}, where performance is fundamentally constrained by grounding limitations. However, these benchmarks largely emphasize complex instruction following rather than deliberative, multi-step planning, and thus fail to rigorously assess LLMs' ability to verify plans and backtrack under global resource constraints.

\subsection{Evaluation of LLM Planning}
In the era of LLMs, planning has been widely studied as a high-level reasoning capability central to achieving human-level intelligence. Early studies on classical~\citep{valmeekam2023planbench,stechly2025on} and temporal planning tasks~\citep{zhang-etal-2024-timearena,ding-etal-2025-tcp} show that LLM-based agents struggle to plan reliably under specific global constraints. Recent benchmarks shift to real-world scenarios like complex travel planning~\citep{xie2024travelplanner,shao2025chinatravel,qu2025tripscore,singh-etal-2024-personal,wang-etal-2025-triptailor}, but often simplify planning horizons (e.g., day-level instead of minute-level)~\citep{xie2024travelplanner,qu2025tripscore} or lack complex constraints~\citep{shao2025chinatravel}. These limitations motivate the need for a more comprehensive benchmark to evaluate long-horizon agent planning.

\begin{figure*}[!t]
\centering
\includegraphics[width=0.95\linewidth]{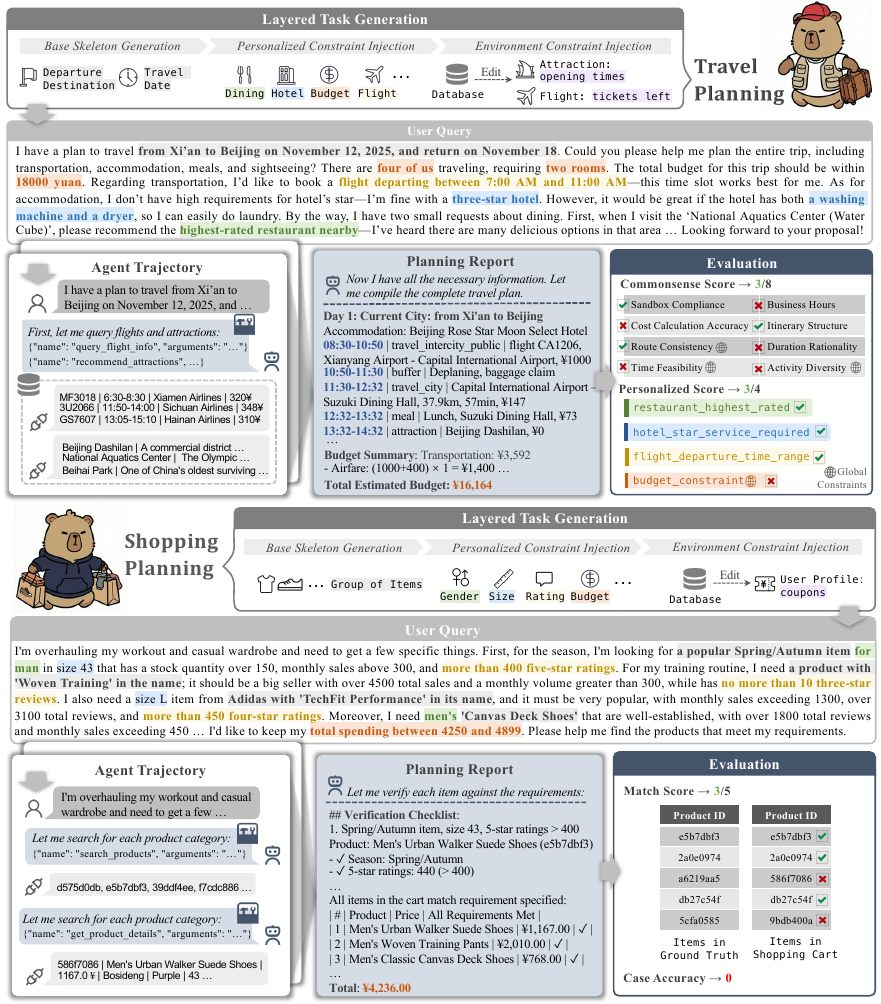}
\caption{Overview of the proposed benchmark {\benchmark}.}
\label{fig:framework}
\end{figure*}

\section{{\benchmark}}
\subsection{Benchmark Overview}
We illustrate {\benchmark} in Figure~\ref{fig:framework}, which systematically evaluates language agents across two complex, real-world scenarios: \textit{Travel Planning} and \textit{Shopping Planning}. Each task runs in a self-contained offline sandbox backed by a database, which is accessible only via the provided Python toolkits, ensuring reproducibility and ease of deployment. During execution, an agent receives a user query, iteratively invokes tools to retrieve the required information, and produces a plan that satisfies both task policies and user needs.

Concretely, \textit{Travel Planning} consists of 120 unique tasks, each available in both Chinese and English. Every task runs in an isolated sandbox with seven task-specific sub-databases accessed via 9 specialized APIs (e.g., \texttt{query\_hotel\_info}). Agents must generate realistic multi-day itineraries by integrating information about transportation, accommodation, attractions, and dining, producing a structured, minute-level schedule with itemized costs and a final budget summary. The core difficulty lies in the tight coupling among time, location, and budget, where small changes can cascade into downstream failures. \textit{Shopping Planning} comprises 120 English tasks, each running in an isolated sandbox with three sub-databases and 15 specialized APIs (e.g., \texttt{search\_products}). Agents must construct an optimal purchasing plan by combining user requirements with product details, sizing, shipping times, coupon availability, and budget constraints (see Appendix~\ref{app:Travel Planning Output Example}). Using these tools, they assemble a final shopping cart returned as a structured JSON object. Agents must construct an optimal purchasing plan by combining user requirements with product specifications, sizing, shipping times, coupon availability, and budget constraints, ultimately returning a structured JSON shopping cart (see Appendix~\ref{app:Shopping Output Example}). Challenges lies in the combinatorial optimization across user needs, product attributes, coupon applicability, and budget.

Table~\ref{tab:benchmark_statistics} summarizes key statistics for each domain in {\benchmark}, along with the aspects assessed during evaluation. We describe the goals and challenges in terms of the three core capabilities required of capable agents later.

\begin{table}[!t]
\centering
\footnotesize
\caption{Key statistics and evaluation subjects for {\benchmark}.}
\scalebox{0.8}{
\begin{tabular}{l c c}
\toprule
 & \textbf{Travel} & \textbf{Shopping} \\
\midrule
\# Tasks (Language) & 120 (ZH) / 120 (EN) & 120 (EN) \\
\# Available APIs & 9 & 15 \\
\# Avg. Records per Task & 7,708 & 171 \\
\midrule
\multirow{2}{*}{Evaluation Subject} & Structured & Shopping \\
 & itinerary report & list \\
\bottomrule
\end{tabular}
}
\label{tab:benchmark_statistics}
\end{table}

\paragraph{Proactive Information Acquisition}
In \textit{Travel Planning}, agents must discover key environment states via APIs rather than making facts in a plan. A common failure is planning intra-city travel (\texttt{travel\_city}), which requires a complex, multi-step query (e.g., location coordinates $\rightarrow$ available transportation options) that is easy to overlook. In \textit{Shopping Planning}, beyond the information explicitly stated in the user query, agents must uncover critical implicit details. For example, user-specific attributes such as clothing size or shipping destination are often omitted and must be actively retrieved via the appropriate tools.

\paragraph{Local Constrained Reasoning}
In \textit{Travel Planning}, agents must handle both explicit user preferences stated in the query (e.g., a three-star hotel with a washing machine) and implicit constraints that emerge during interactions with the environment (e.g., a target attraction closed today). 
In \textit{Shopping Planning}, agents must satisfy explicit user preferences by selecting appropriate tools and applying relevant filters, such as preferred brands, colors, or minimum required product ratings. They must integrate these multi-faceted constraints to identify the product that best matches the user's intent.

\paragraph{Global Constrained Optimization}
In \textit{Travel Planning}, the itinerary must satisfy global constraints across three dimensions: \textbf{time} (avoid overlaps; respect visit durations and opening hours), \textbf{space} (ensure feasible transportation between consecutive locations), and \textbf{budget} (accurately aggregate costs and comply with trip-level limits). 
In \textit{Shopping Planning}, the challenge extends beyond selecting individual items to managing combinatorial constraints across the entire cart. Agents must assemble a set of products that fits the total budget, often prioritizing the lowest \textbf{overall cart cost} rather than the cheapest items in isolation. Likewise, effective coupon use may require choosing a higher-priced item to reduce the final total.

\subsection{Benchmark Construction}
A three-stage data construction pipeline is utilized to generate high-complexity agent planning tasks along with verifiable, unique solutions. Figure \ref{fig:framework} highlights the central stage, \textit{Layered Task Generation}, to aid understanding.

\paragraph{Stage 1: Database and Toolbox Design.}
To build sandbox environments for agent planning, we develop domain-specific databases and a suite of custom Python APIs that abstract database query behavior. We populate these databases with domain-specific data. For \textit{Travel Planning}, we use public APIs (e.g., Fliggy, Amap and Web Search) to collect real-world data from popular tourist cities in China, covering transportation, accommodation, dining, and attractions, with key fields such as prices, schedules, geographic coordinates, and ratings. For \textit{Shopping Planning}, we synthesize fine-grained product data to enable controlled complexity, including attributes such as price, stock levels, monthly sales, user ratings, and promotion details. Tools in each domain follow a hierarchical design that encourages multi-step agent interaction with the environment. See Appendix~\ref{app:tool schema} and Appendix~\ref{app:database_schema} for detailed tool schemas and example database entries.

\paragraph{Stage 2: Layered Task Generation.}
We adopt a solution-centric, reverse-generation process to construct complex agent planning tasks by progressively adding constraints:
\begin{itemize}[leftmargin=*]
    \item \textit{Base Skeleton Generation.}
    We derive a base skeleton from the database: in \textit{Travel Planning}, this skeleton specifies the departure city, destination, and travel date; in \textit{Shopping Planning}, it defines a set of items with a common theme (e.g., clothes and shoes for spring travel).
    
    \item \textit{Personalized Constraint Injection.}
    We build domain-specific pools of personalized constraints and sample multiple, potentially complex constraints to augment the task skeleton. In \textit{Travel Planning}, examples include ``\textit{recommend the highest‑rated restaurant nearby}'' or ``\textit{book a flight departing after 7:00 AM}''. In \textit{Shopping Planning}, examples include ``\textit{find a product named \texttt{ShockWave} with a rating above 4.7}'' or ``\textit{keep my total spending above 4500 yuan}''.
    
    \item \textit{Environment Constraint Injection.}
    We further inject implicit environmental constraints that introduce dynamic challenges discoverable only through tool use. In \textit{Travel Planning}, these include cases where a key attraction is closed on the planned day or where flights have limited ticket availability. In \textit{Shopping Planning}, they appear as combinatorial optimization problems, such as coupon-stacking rules that make a seemingly more expensive cart the cheapest after all discounts are applied. After adding these constraints, candidates in databases are automatically adjusted so that exactly one optimal solution exists, ensuring solvability and uniqueness. 
\end{itemize}
Finally, we use an LLM to convert all structured constraints into a conversational user query for each agent task.

\paragraph{Stage 3: Manual Quality Control.}
This stage provides the final quality check for generated tasks. Human experts review and revise the LLM-generated queries and the full task to ensure: (i) natural, fluent language; (ii) clear, unambiguous logic; and (iii) a reachable solution for every task, ensuring the overall quality of {\benchmark}.

\subsection{Task Evaluation}
To ensure accurate and consistent task evaluation in {\benchmark}, we rely on code-based automated evaluation rather than LLM-based scoring. 

\paragraph{Travel Planning.} In \textit{Travel Planning}, we first use Qwen-Plus-2507 to parse the models' natural-language itineraries into a predefined structured format, and then apply the following metrics for rule-based scoring:

\begin{itemize}[label=-, leftmargin=*]
    \item \textit{Commonsense Score.}
    We evaluate agents' ability to produce a real-world feasible plan along eight dimensions spanning 21 checkpoints, including route consistency, sandbox compliance, itinerary structure, time feasibility, business hours compliance, duration rationality, cost calculation accuracy, and activity diversity. All sub-checkpoints are verified automatically via code execution. Each dimension contributes equally to the commonsense score: it receives $1/8$ if all its sub-checkpoints are satisfied, and $0$ otherwise. For each task, the commonsense score is the sum of the eight dimension scores. See Appendix~\ref{app:travel_cs_score} for the full taxonomy and descriptions.
    
    \item \textit{Personalized Score.}
    We use this metric to evaluate agents' ability to satisfy user-specific requirements stated in the query. All injected personalized constraints in the Layered Task Generation stage are automatically translated into code-based checks, which are then used to verify whether the structured plan meets the user's needs. For each task, the personalized score is $1$ if all constraints are satisfied and $0$ otherwise.
    
    \item \textit{Composite Score.}
    This metric provides a holistic evaluation of agent performance on \textit{Travel Planning}. For each task, it is the average of the \textit{Commonsense Score} and the \textit{Personalized Score}.

    \item \textit{Case Accuracy.}
    This metric is a stricter, case-level measure of performance. For each task, it is $1$ only if both the \textit{Commonsense Score} and the \textit{Personalized Score} are perfect, and $0$ otherwise.
\end{itemize}

\begin{table*}[!t]
    \centering
    \caption{Evaluation results of agentic LLMs on {\benchmark}. CS, PS, and Comp denote \textit{Commonsense}, \textit{Personalized}, and \textit{Composite} Scores, respectively. \textbf{Avg Acc.} is the mean Case Accuracy across both domains. Results are averaged over four runs. We \textbf{bold} the best and \underline{underline} the second-best result in each group.}
    \label{tab:main_results}

    \small
    \setlength{\tabcolsep}{3.0pt}
    \scalebox{0.9}{
    \renewcommand{\arraystretch}{0.95}
    \begin{tabular}{l >{\columncolor{violet!5}}c cccccc}
        \toprule
        \multirow{2}{*}{\textbf{Model}} &
        \multirow{2}{*}{\textbf{Avg Acc.}} &
        \multicolumn{4}{c}{\textbf{Travel Planning}} &
        \multicolumn{2}{c}{\textbf{Shopping Planning}} \\
        \cmidrule(lr){3-6} \cmidrule(lr){7-8}
        &  & CS Score & PS Score & Comp Score & Case Acc.
        & Match Score & Case Acc. \\
        \midrule

        \rowcolor{tablegroupcolor}
        \multicolumn{8}{c}{\textit{Non-Reasoning Models}} \\
        Anthropic/Claude-4.5-Opus (w/o thinking) & \textbf{26.3} & \textbf{67.5} & \textbf{58.8} & \textbf{63.1} & \textbf{6.7} & \textbf{82.2} & \textbf{45.8} \\
        Anthropic/Claude-4.5-Sonnet (w/o thinking) & \underline{17.2} & 53.4 & 42.8 & \underline{48.1} & \underline{1.1} & \underline{75.8} & \underline{33.3} \\
        Alibaba/Qwen3-Max (w/o thinking) & 12.8 & 36.7 & 30.7 & 31.8 & 0.8 & 70.2 & 24.7 \\
        ByteDance/Seed-1.8-minimal & 11.3 & 43.0 & \underline{47.5} & 45.3 & 0.0 & 68.1 & 22.5 \\
        Alibaba/Qwen-Plus (w/o thinking) & 7.5 & 37.3 & 13.0 & 25.1 & 0.0 & 63.9 & 15.0 \\
        Z.ai/GLM-4.7 (w/o thinking) & 7.1 & 38.9 & 22.5 & 30.7 & 0.0 & 61.2 & 14.2 \\
        DeepSeek-AI/DeepSeek-V3.2 (w/o thinking) & 5.3 & 37.4 & 12.1 & 24.7 & 0.0 & 58.3 & 10.6 \\
        OpenAI/GPT-5.2-none & 4.5 & \underline{54.3} & 29.9 & 42.1 & 0.4 & 58.6 & 8.6 \\
        xAI/Grok-4.1-Fast (non-reasoning) & 3.0 & 39.6 & 19.7 & 29.6 & 0.0 & 50.1 & 5.9 \\
        \midrule

        \rowcolor{tablegroupcolor}
        \multicolumn{8}{c}{\textit{Reasoning Models}} \\
        OpenAI/GPT-5.2-high & \textbf{44.6} & \textbf{88.5} & \textbf{83.3} & \textbf{85.8} & \textbf{35.0} & \textbf{84.8} & \textbf{54.2} \\
        Anthropic/Claude-4.5-Opus (w/ thinking) & \underline{33.9} & \underline{79.3} & \underline{70.9} & \underline{75.1} & \underline{22.7} & 80.0 & 45.0 \\
        OpenAI/GPT-5-high & 31.6 & 78.7 & 65.9 & 72.3 & 18.9 & 80.4 & 44.2 \\
        Google/Gemini-3-Flash-Preview & 28.8 & 67.1 & 57.7 & 62.4 & 5.9 & 80.6 & \underline{51.7} \\
        Alibaba/Qwen3-Max (w/ thinking) & 28.7 & 64.0 & 61.7 & 62.8 & 13.8 & \underline{82.6} & 43.5 \\
        Anthropic/Claude-4.5-Sonnet (w/ thinking) & 25.5 & 65.2 & 58.4 & 61.8 & 7.6 & 80.0 & 43.3 \\
        OpenAI/o3 & 24.9 & 76.5 & 55.6 & 66.1 & 11.3 & 76.9 & 38.5 \\
        Google/Gemini-3-Pro-Preview & 23.2 & 58.4 & 25.1 & 41.8 & 0.7 & 78.0 & 45.8 \\
        Deepseek-AI/DeepSeek-V3.2 (w/ thinking) & 21.6 & 47.4 & 35.0 & 41.2 & 0.7 & 78.8 & 42.5 \\
        ByteDance/Seed-1.8-high & 20.4 & 43.6 & 56.7 & 50.1 & 0.0 & 77.5 & 40.8 \\
        xAI/Grok-4.1-Fast (reasoning) & 17.2 & 57.1 & 37.7 & 47.4 & 2.7 & 74.0 & 31.7 \\
        Alibaba/Qwen-Plus (w/ thinking) & 17.1 & 35.4 & 22.4 & 28.9 & 0.0 & 73.3 & 34.1 \\
        Google/Gemini-2.5-Pro & 17.0 & 62.3 & 42.0 & 52.2 & 3.2 & 69.1 & 30.8 \\
        Z.ai/GLM-4.7 (w/ thinking) & 14.0 & 44.0 & 44.6 & 44.3 & 0.4 & 72.5 & 27.5 \\
        OpenAI/o4-mini & 12.4 & 58.0 & 36.6 & 47.2 & 3.0 & 69.1 & 21.7 \\
        Moonshot-AI/Kimi-K2-Thinking & 12.1 & 45.2 & 32.5 & 38.9 & 0.0 & 65.8 & 24.2 \\
        \bottomrule
    \end{tabular}
    }
\end{table*}

\paragraph{Shopping Planning.} We compare the products the agent adds to the shopping cart against the ground-truth cart and evaluate their performance using the following metrics:
\begin{itemize}[label=-, leftmargin=*]
    \item \textit{Match Score.}
    This metric measures the agent's ability to identify the products requested by the user. For each task, it is computed as the number of products in the agent's cart that match the ground-truth items, divided by the total number of ground-truth items.

    \item \textit{Case Accuracy.}
    This metric provides a stricter measure of performance. For each task, it is $1$ only if all products in the cart exactly match the ground-truth products and $0$ otherwise.
\end{itemize}

\section{Experiments}
\subsection{Experimental Setup}
We benchmark a wide range of state-of-the-art LLMs on {\benchmark}, including OpenAI GPT and o series~\citep{openai_gpt5,openai_o3}, Anthropic Claude series~\citep{anthropic_claude45_opus}, Google Gemini series~\citep{comanici2025gemini,google_gemini3_pro}, xAI Grok series~\citep{xai_grok4}, DeepSeek-AI DeepSeek series~\citep{liu2025deepseek}, Alibaba Qwen series~\citep{yang2025qwen3}, Z.ai GLM series~\citep{zeng2025glm}, Moonshot-AI Kimi  series~\citep{team2025kimi}, ByteDance Seed series~\citep{seed2025seed1}. 
We experiment on both reasoning and non-reasoning models. For hybrid-thinking model families, the reasoning effort is set to the maximum when operating in thinking mode. All models are instantiated as function-calling agents, with tools specified in the OpenAI tool schema format, and a maximum of 400 tool calls allowed per task. For robustness, each task is run four times, and results are averaged across all runs and tasks. Notably, results for \textit{Travel Planning} are averaged over both the Chinese and English case variants. Detailed system prompts for each domain are provided in Appendix~\ref{app:prompts}.

\subsection{Main Results}
Table~\ref{tab:main_results} presents the evaluation results of all models on {\benchmark}. We rank the models by their average case accuracy across both domains, revealing several notable findings:

\paragraph{Planning Fragility of LLM Agents.}
Even frontier LLM agents are unable to fully solve complex planning tasks in {\benchmark}. A clear discrepancy emerges between relatively high constraint-level scores and low case-level accuracy. In \textit{Travel Planning}, for example, even the best-performing model produces fully correct plans in only 35.0\% of cases. Although a model may satisfy most individual requirements and achieve a high composite score, a single critical failure---such as exceeding the budget or introducing a scheduling conflict---can invalidate the entire plan. This fragility is also evident in \textit{Shopping Planning}, where high Match Scores frequently fail to translate into high Case Accuracy. These results suggest that current agents struggle to integrate individually correct actions into a globally coherent and executable plan, exposing fundamental limitations in end-to-end long-horizon planning.

\paragraph{Performance Gains from Internal Reasoning.}
In general, the benchmark is led by models from the GPT-5 series and Claude-4.5 when operated with maximal reasoning effort. Notably, we observe pronounced domain specialization across models: although Gemini-3-Flash-Preview performs poorly on \textit{Travel Planning}, it excels in \textit{Shopping Planning}, outperforming all other baselines with a Case Accuracy of 60.0\%. Crucially, models equipped with deliberate internal reasoning consistently surpass their non-reasoning counterparts, underscoring the importance of such capabilities for long-horizon planning tasks.

\subsection{Cost--Performance Trade-offs}
\begin{figure}[!t]
    \centering
    \includegraphics[width=\linewidth]{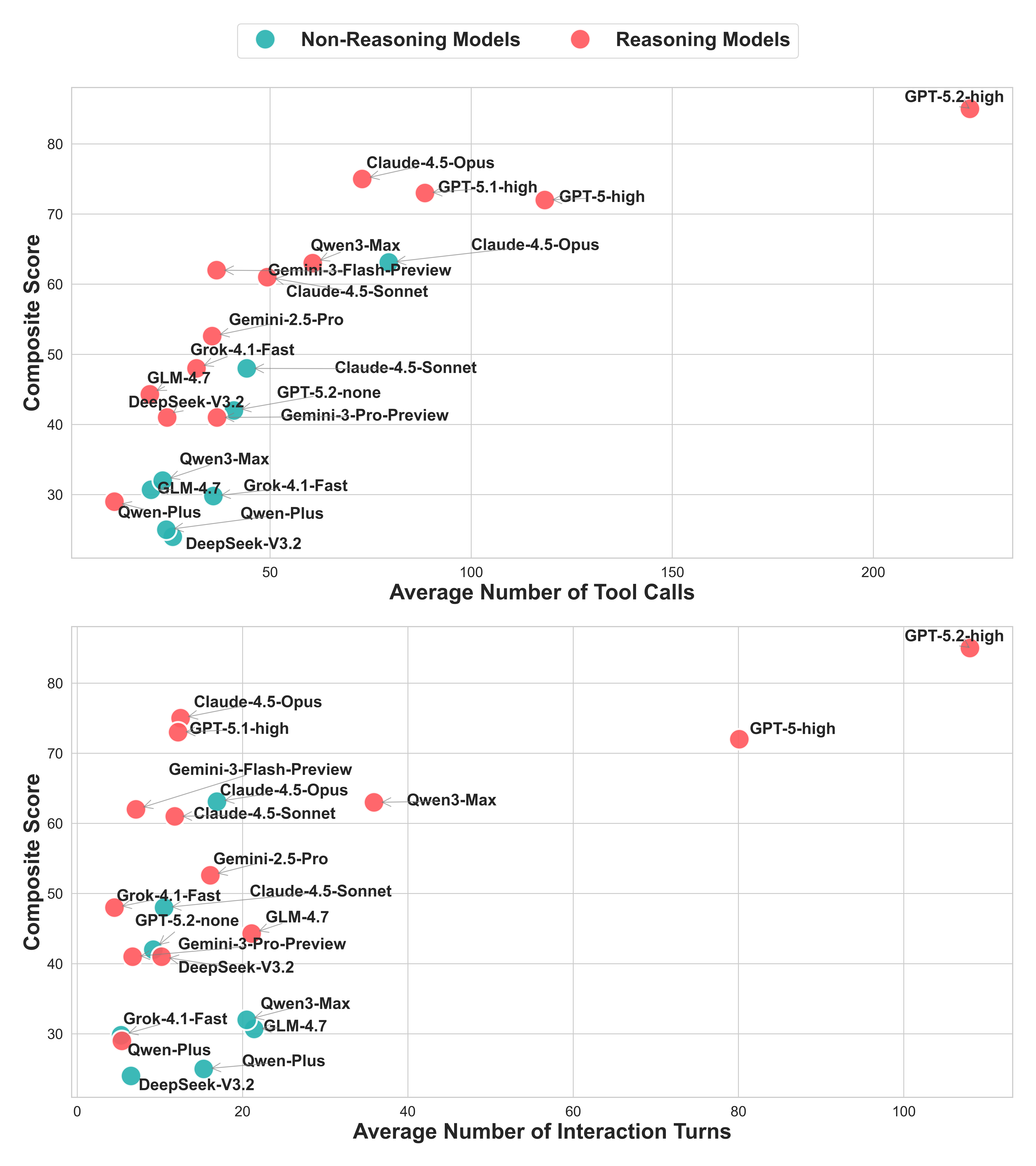}
    \caption{Model performance versus task execution cost on \textit{Travel Planning}. Performance is calculated across all tasks, while cost is measured by the average number of tool calls per task (top) and the average number of interaction turns per task (bottom).}
    \label{fig:toolcalls_turns_vertical}
\end{figure}

We further examine the relationship between inference cost and agent performance on {\benchmark}. Using \textit{Travel Planning} as an example, Figure~\ref{fig:toolcalls_turns_vertical} plots the average composite score across all tasks against agent cost, measured by the average number of tool invocations and interaction turns per task. We observe the following:

\paragraph{More Tool Use Yields Higher Performance.}
As shown at the top of Figure~\ref{fig:toolcalls_turns_vertical}, model performances generally increase with the number of tool calls, suggesting that long-horizon agent planning in real-world tasks relies on extensive, proactive information gathering (e.g., validating times, checking transfer feasibility, and verifying budgets) beyond parametric knowledge. In extreme cases, GPT-5.2-high achieves the best score while making roughly 224 tool invocations per task.

\paragraph{Reasoning Improves the Efficiency Frontier.}
Reasoning models consistently sit in a better part of the trade-off curve, achieving higher scores with comparable or lower interaction costs. For example, enabling thinking mode in Claude-4.5-Opus both improves performance and reduces interaction turns (16.9$\rightarrow$12.5) and tool calls (79.5$\rightarrow$72.9). This suggests that internal planning curbs redundant trial-and-error and consolidates tool use into fewer, higher-quality actions.

\paragraph{Interaction Patterns: Sequential vs. Parallel Execution.}
Models exhibit distinct interaction strategies even within the same lineage (Figure~\ref{fig:toolcalls_turns_vertical}). Within the GPT-5 family, GPT-5.1-high favors parallel execution by bundling multiple tool calls into a single turn, whereas GPT-5.2-high adopts a more sequential, step-by-step workflow. Although GPT-5.2-high achieves better performance (+12.7\%), it requires nearly $10\times$ more turns than GPT-5.1-high, highlighting a clear cost-performance trade-off between parallel, more efficient interaction and sequential thorough verification.

\subsection{Impact of Task Complexity}
\begin{figure}[!t]
    \centering
    \includegraphics[width= \linewidth]{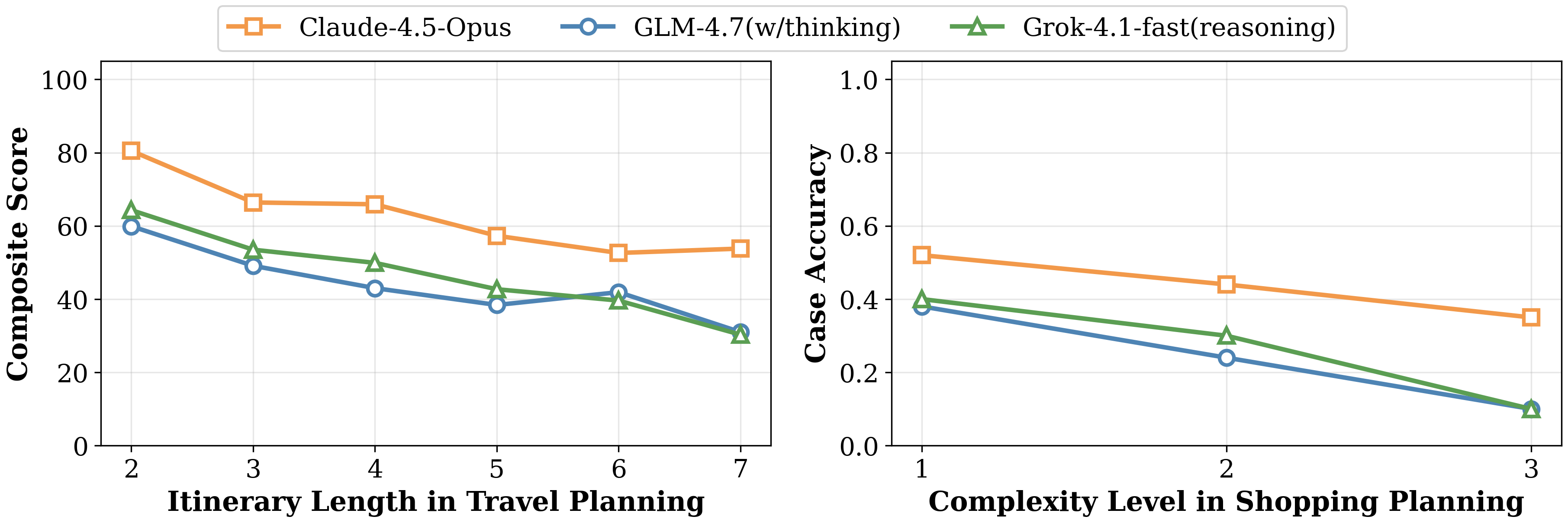}
    \caption{Model performance versus task complexity in {\benchmark}. In each domain, performance is calculated across tasks at each complexity level.}
    \label{fig:difficulty_level}
\end{figure}

We investigate how task complexity influences model performance in {\benchmark}. We report results for three models from distinct families: Claude-4.5-Opus, GLM-4.7 (w/ thinking), and Grok-4.1-Fast (reasoning). As shown in Figure~\ref{fig:difficulty_level}, model performance consistently drops as task complexity increases in both domains. In \textit{Travel Planning}, composite scores decline as itinerary length increases from 2 to 7 days, indicating that long-horizon planning remains a major bottleneck. As the number of days grows, agents must retrieve and integrate substantially more information, increasing runtime and expanding the space of interacting constraints. Consequently, small local mistakes can propagate across days and ultimately render the overall itinerary infeasible. In \textit{Shopping Planning}, case accuracy declines from Level 1 to Level 3 as tasks incorporate more cross-item constraints (e.g., price-range requirements for Level 2, and coupon timing for Level 3). This progression shifts the problem from straightforward item matching to a global, joint optimization challenge.

\subsection{Error Pattern Analysis}
\begin{figure}[!t]
  \centering
  \includegraphics[width=1.0\columnwidth]{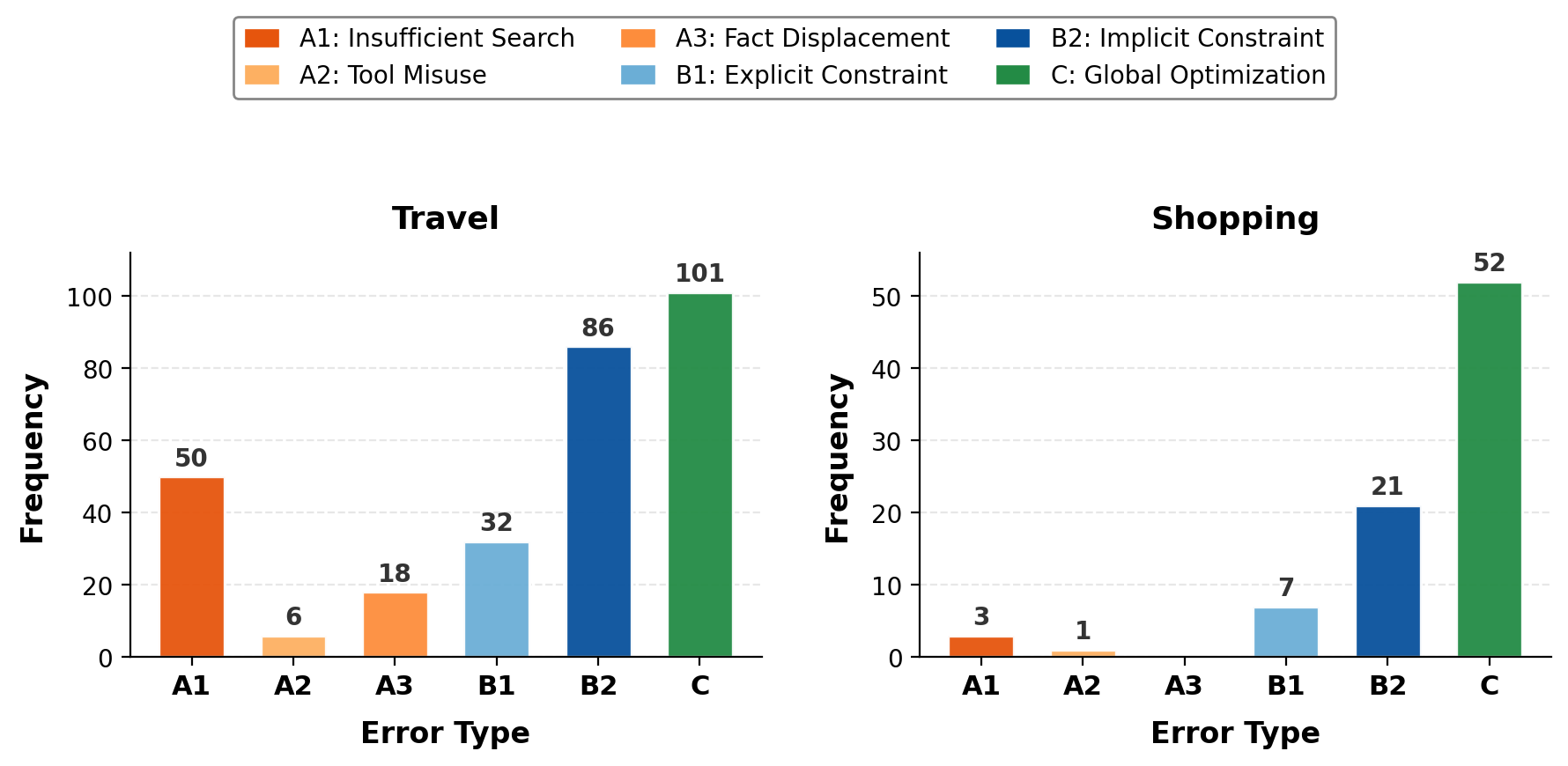}
  \caption{Error pattern distribution of Claude-4.5-Opus (with reasoning enabled) on {\benchmark}.}
  \label{fig:error_distribution}
\end{figure}

To better understand what drives agent failures in {\benchmark}, we categorize error patterns into the agent competencies discussed earlier and annotate 140 failed trajectories generated by Claude-4.5-Opus (w/ thinking): 80 from \textit{Travel Planning} and 60 from \textit{Shopping Planning}. Note that a single failed trajectory may result from a combination of issues. We show their distribution in Figure~\ref{fig:error_distribution}.

\paragraph{Pattern A: Information Acquisition Failures.}
This category covers failures in retrieving, perceiving, or correctly using essential environmental information. \textbf{A1: Insufficient Search} refers to the omission of querying critical information. In \textit{Travel Planning}, this error often exists as agents overlook transit times or distances between specific locations; moreover, as the number of required attractions increases, agents are more likely to skip these critical search steps. \textbf{A2: Tool Misuse} involves selecting inappropriate tools or providing malformed arguments (e.g., coordinate precision mismatches). \textbf{A3: Fact Displacement} arises when agents retrieve the correct information but later misstate it in the final plan or replace it with fabricated values (e.g., recording a retrieved \textyen 100 price as \textyen 150 in the itinerary).

\paragraph{Pattern B: Local Reasoning Failures.}
These errors arise when the model fails to satisfy constraints at specific decision points despite having access to the correct information. \textbf{B1: Explicit Constraint} violations disregard user-specified requirements (e.g., ignoring a ``three-star hotel'' preference). \textbf{B2: Implicit Constraint} (86 in Travel, 21 in Shopping) occurs when plans conflict with common sense or environmental reality (e.g., attempting to book flights for 4 people when only 2 tickets are available), indicating deficiencies in non-explicit constraint reasoning.

\paragraph{Pattern C: Global Optimization Failures.}
As the most prevalent failure (101 in \textit{Travel}, 52 in \textit{Shopping}), this category reflects a breakdown in integrating interdependent decisions into a holistic plan. This involves an inability to perform systematic trade-offs under overarching constraints, such as failing to find the optimal combination of items and coupons to achieve the lowest final price in shopping, or violating diversity requirements in travel (e.g., repetitive restaurant types or attraction categories over a multi-day trip). It also covers structural incoherence, such as temporal overlaps between steps or logical discontinuities between days, indicating that agents struggle with combinatorial optimization and with satisfying multiple complex constraints in long-horizon tasks.

\section{Conclusion}

We introduce \textsc{\benchmark}, a challenging benchmark for assessing long-horizon LLM agent planning in real-world settings. Building on multi-day travel and complex shopping scenarios, it enables a systematic evaluation of proactive information gathering and multi-level constrained optimization. Our analysis shows that even state-of-the-art agents struggle significantly with these tasks, revealing a fundamental gap in their planning reliability. This work provides a foundation for developing next-generation agents capable of tackling complex, grounded planning challenges.


\section*{Limitations}
Though carefully designed, our proposed benchmark still has several limitations. First, the agent-planning tasks and corresponding environments in {\benchmark} are restricted to the travel and shopping domains. Incorporating a broader range of real-world scenarios would make the benchmark more comprehensive. Second, although we use real-world data to construct databases in sandboxes, the user queries are synthesized by adding multi-level constraints, which may lead to a distribution shift relative to real user queries in these domains. Finally, while the tasks in this work focus on single-turn, multi-step agent planning, modeling multi-turn user-agent interactions remains an important direction for future work.

\bibliography{custom,anthology-1,anthology-2}

\clearpage

\appendix
\section{Example of Task Outputs}
\label{app:task_output_example}

\subsection{Travel Planning Output Example}
\label{app:Travel Planning Output Example}
Box~\ref{prompt:Travel Planning Output Example} shows an example output for the travel planning task: a two-day travel plan including the itinerary and budget.

\subsection{Shopping Output Example}
\label{app:Shopping Output Example}
Box~\ref{prompt:Example for cart checkout} shows an example output for the shopping planning task: a shopping cart that includes the items selected by the agent and the coupons to be used.

\section{Tool Schemas and Database Schema}
\label{app:tools_and_db}
\subsection{Tool Schema}
\label{app:tool schema}
Table \ref{tab:travel_tools} presents the definitions of the tools used in the travel planning task. Table \ref{tab:shopping_tools} presents the definitions of the tools used in the shopping planning task.

\subsection{Database Schema}
\label{app:database_schema}
Table \ref{tab:Travel Planning_db_schema} lists the detailed database fields in the travel planning task and explains their meanings. Table \ref{tab:Shopping Planning_db_schema} lists the detailed database fields in the travel planning task and explains their meanings.

\section{Evaluation Details}
\label{app:travel_cs_score}
Table \ref{tab:travel_commonsense} shows the detailed descriptions of the 21 scoring criteria for commonsense score in Travel Planning.

\section{Prompt Template}
\label{app:prompts}
This section presents the system prompts for the Travel Planning (Box \ref{app:travel_system_prompt}) and Shopping Planning tasks (Box \ref{app:shop_system_prompt}), as well as the prompt used for plan format conversion in the Travel Planning domain (Box \ref{app:Plan Format Conversion Prompt}).

\onecolumn

\begin{promptbox}[label={prompt:Travel Planning Output Example}]{Travel Planning Output Example}
Day 1:
Current City: from Hefei to Nanjing
Accommodation: Orange Hotel Nanjing Confucius Temple Scenic Area, 441RMB/room/night
06:19-07:14 | travel_intercity_public | train G7798, Hefei Station - Nanjing South Station, 67RMB/person
07:14-07:44 | buffer | Disembark and exit station
07:44-07:56 | travel_city | Nanjing South Station - Orange Hotel Nanjing Confucius Temple Scenic Area, 8.3km, 12min, 31RMB
07:56-08:30 | hotel | Check-in, Orange Hotel Nanjing Confucius Temple Scenic Area
08:30-08:42 | travel_city | Orange Hotel Nanjing Confucius Temple Scenic Area - Nanjing City Wall Taicheng Scenic Area, 7.8km, 12min, 29RMB
08:42-10:30 | attraction | Nanjing City Wall Taicheng Scenic Area, 30RMB/person
10:30-10:38 | travel_city | Nanjing City Wall Taicheng Scenic Area - Nanjing Deji Plaza, 5.4km, 8min, 22RMB
10:38-12:00 | attraction | Nanjing Deji Plaza, 0RMB/person
12:00-12:05 | travel_city | Nanjing Deji Plaza - Nice Meeting You (Deji Plaza Branch), 40m, 5min, 0RMB
12:05-13:15 | meal | Lunch, Nice Meeting You (Deji Plaza Branch), 99RMB/person
13:15-13:23 | travel_city | Nice Meeting You (Deji Plaza Branch) - Laomendong, 5.1km, 8min, 21RMB
13:23-16:00 | attraction | Laomendong, 0RMB/person
16:00-17:00 | buffer | Rest and explore Laomendong area
17:00-18:30 | meal | Dinner (Birthday Celebration), Six Dynasties Pine Teahouse, 294RMB/person
18:30-18:53 | travel_city | Six Dynasties Pine Teahouse - Orange Hotel Nanjing Confucius Temple Scenic Area, 1.9km, 23min, 0RMB
18:53-24:00 | hotel | Rest, Orange Hotel Nanjing Confucius Temple Scenic Area

Day 2:
Current City: from Nanjing to Hefei
Accommodation: -
08:00-08:30 | hotel | Check-out, Orange Hotel Nanjing Confucius Temple Scenic Area
08:30-08:37 | travel_city | Orange Hotel Nanjing Confucius Temple Scenic Area - Nanjing Museum, 5km, 7min, 21RMB
08:37-09:00 | buffer | Wait for museum to open
09:00-12:30 | attraction | Nanjing Museum, 0RMB/person
12:30-12:35 | travel_city | Nanjing Museum - Yichun Restaurant (Nanjing Museum Branch), 240m, 5min, 0RMB
12:35-13:45 | meal | Lunch, Yichun Restaurant (Nanjing Museum Branch), 53RMB/person
13:45-13:57 | travel_city | Yichun Restaurant (Nanjing Museum Branch) - Lion Bridge Pedestrian Street, 7.7km, 12min, 29RMB
13:57-16:00 | attraction | Lion Bridge Pedestrian Street, 0RMB/person
16:00-16:23 | travel_city | Lion Bridge Pedestrian Street - Nanjing South Station, 15.7km, 23min, 58RMB
16:23-17:18 | buffer | Waiting at station
17:18-17:48 | buffer | Security check and boarding
17:48-18:39 | travel_intercity_public | train G3031, Nanjing South Station - Hefei Station, 67RMB/person

**Budget Summary**:

**Transportation: 613RMB**
- Train: (67RMB + 67RMB) * 3 persons = 402RMB
- City Transport: 31RMB + 29RMB + 22RMB + 21RMB + 21RMB + 29RMB + 58RMB = 211RMB (1 vehicle is sufficient for 3 people)

**Accommodation: 882RMB**
- Orange Hotel Nanjing Confucius Temple Scenic Area: 441RMB * 2 rooms * 1 night = 882RMB

**Meals: 1,338RMB**
- Day 1 Lunch (Nice Meeting You): 99RMB * 3 = 297RMB
- Day 1 Birthday Dinner (Six Dynasties Pine Teahouse): 294RMB * 3 = 882RMB
- Day 2 Lunch (Yichun Restaurant): 53RMB * 3 = 159RMB

**Attractions \& Tickets: 90RMB**
- Nanjing City Wall Taicheng Scenic Area: 30RMB * 3 = 90RMB
- Nanjing Deji Plaza: Free
- Laomendong: Free
- Nanjing Museum: Free
- Lion Bridge Pedestrian Street: Free

**Total Estimated Budget: 2,923RMB**

\end{promptbox}

\begin{promptbox}[label={prompt:Example for cart checkout}]{Example for cart checkout}

{
    "user_id": "6fefe14b",
    "user_name": "SophieChen",
    "items": [
        {
            "product_id": "aceae063",
            "name": "Nike Air Zoom Pegasus - Women's Running Shoes",
            "quantity": 1,
            "price": 899.0
        },
        {
            "product_id": "dd7e1db1",
            "name": "Women's 'Cloud-Feel' Performance Long-Sleeve Top",
            "quantity": 1,
            "price": 289.0
        },
        {
            "product_id": "bfb9b6e8",
            "name": "Women's Summer Linen Blend Wide-Leg Trousers",
            "quantity": 1,
            "price": 899.0
        },
        {
            "product_id": "798abb6c",
            "name": "Women's Courtright Canvas Sneakers",
            "quantity": 1,
            "price": 549.9
        },
        {
            "product_id": "2239e57b",
            "name": "Gucci Women's Princetown Linen Slippers",
            "quantity": 1,
            "price": 6850.0
        },
        {
            "product_id": "0fc1af7f",
            "name": "Women's Armour-Fit Training Leggings",
            "quantity": 1,
            "price": 429.0
        }
    ],
    "used_coupons": [],
    "summary": {
        "total_items_count": 6,
        "total_price": 9915.9
    }
}
\end{promptbox}

\clearpage

\begin{table*}[ht]
\centering
\caption{Tool schemas designed for \textit{Travel Planning} scenario in {\benchmark}.}
\label{tab:travel_tools}
\resizebox{\textwidth}{!}{%
\begin{tabular}{@{}l p{5.5cm} l@{}}
\toprule
\textbf{Tool} & \textbf{Tool Description} & \textbf{Parameters} \\
\midrule
\texttt{query\_train\_info} & 
Query train ticket information by origin, destination, date and other conditions. Returns train number, departure/arrival time, station information, journey duration, seat class, remaining seats status, price, etc. & 
\begin{tabular}[t]{@{}l@{}}
    \textbf{origin} (\textit{string, required}): Origin city name. \\ 
    \textbf{destination} (\textit{string, required}): Destination city name. \\ 
    \textbf{depDate} (\textit{string, required}): Departure date (YYYY-MM-DD). \\ 
    \textbf{seatClassName} (\textit{string}): Seat class (First/Second/Business).
\end{tabular} \\
\addlinespace

\texttt{query\_flight\_info} & 
Query flight information by origin, destination, date and other conditions. Returns flight number, departure/arrival time, airport information, journey duration, seat class, price, aircraft type, etc. & 
\begin{tabular}[t]{@{}l@{}}
    \textbf{origin} (\textit{string, required}): Origin city name. \\ 
    \textbf{destination} (\textit{string, required}): Destination city name. \\ 
    \textbf{depDate} (\textit{string, required}): Departure date (YYYY-MM-DD). \\ 
    \textbf{seatClassName} (\textit{string}): Seat class (Economy/Business/First).
\end{tabular} \\
\addlinespace

\texttt{query\_hotel\_info} & 
Search hotels based on destination, check-in/out dates, star rating, and brand. Returns hotel information matching the criteria. & 
\begin{tabular}[t]{@{}l@{}}
    \textbf{destination} (\textit{string, required}): Destination city/district name. \\ 
    \textbf{checkinDate} (\textit{string, required}): Check-in date (yyyy-MM-dd). \\ 
    \textbf{checkoutDate} (\textit{string, required}): Check-out date (yyyy-MM-dd). \\ 
    \textbf{hotelStar} (\textit{string}): Hotel star rating (1-5). \\ 
    \textbf{hotelBrands} (\textit{string}): Hotel brand name.
\end{tabular} \\
\addlinespace

\texttt{query\_road\_route\_info} & 
Calculate road route details between two locations including distance (meters), duration (minutes) and cost. Auto-selects walking or driving mode. & 
\begin{tabular}[t]{@{}l@{}}
    \textbf{origin} (\textit{string, required}): Origin coordinates (latitude,longitude). \\ 
    \textbf{destination} (\textit{string, required}): Destination coordinates (lat,lng).
\end{tabular} \\
\addlinespace

\texttt{query\_attraction\_details} & 
Query detailed information about an attraction including ID, name, coordinates, description, rating, opening hours, ticket price, and recommended visit duration. & 
\textbf{attraction\_name} (\textit{string, required}): Attraction name. \\
\addlinespace

\texttt{query\_restaurant\_details} & 
Query detailed information about a restaurant including name, coordinates, price per person, opening/closing time, and rating. & 
\textbf{restaurant\_name} (\textit{string, required}): Restaurant name. \\

\texttt{recommend\_attractions} & 
Search and return popular attraction information for a city. Returns attraction name, description, type, etc. based on city keyword. & 
\begin{tabular}[t]{@{}l@{}}
    \textbf{city} (\textit{string, required}): City name or keyword. \\ 
    \textbf{attraction\_type} (\textit{string}): Type filter (Historical/Natural/ \\ Art/Landmark/Theme Park/Leisure).
\end{tabular} \\
\addlinespace

\texttt{recommend\_restaurants} & 
Recommend nearby restaurants based on specified coordinates. Returns restaurant name, coordinates, price per person, rating, business hours, etc. & 
\begin{tabular}[t]{@{}l@{}}
    \textbf{latitude} (\textit{string, required}): Latitude coordinate (6 decimals). \\ 
    \textbf{longitude} (\textit{string, required}): Longitude coordinate (6 decimals).
\end{tabular} \\
\addlinespace

\texttt{search\_location} & 
Query the latitude and longitude coordinates corresponding to a place name, with precision retained to six decimal places. & 
\textbf{place\_name} (\textit{string, required}): Place name to query. \\
\addlinespace

\bottomrule
\end{tabular}%
}
\end{table*}
\clearpage 

\begin{table*}[!ht]
\centering
\caption{Tool schemas designed for \textit{Shopping Planning} scenario in {\benchmark}.}
\label{tab:shopping_tools}
\resizebox{\textwidth}{!}{%
\begin{tabular}{@{}l p{5.5cm} l@{}}
\toprule
\textbf{Tool} & \textbf{Tool Description} & \textbf{Parameters} \\
\midrule
\texttt{search\_products} & 
Handles open-ended queries by performing a semantic search on product information to retrieve an initial set of relevant items. & 
\begin{tabular}[t]{@{}l@{}}
    \textbf{query} (\textit{string, required}): User's natural language query. \\ 
    \textbf{limit} (\textit{integer}): Limits the number of products returned.
\end{tabular} \\
\addlinespace

\texttt{filter\_by\_brand} & 
Filters a list of products by one or more brand names using OR logic. & 
\begin{tabular}[t]{@{}l@{}}
    \textbf{product\_ids} (\textit{array}): List of product IDs to filter. \\ 
    \textbf{brand\_names} (\textit{array, required}): Brand names to match.
\end{tabular} \\
\addlinespace

\texttt{filter\_by\_color} & 
Filters a list of products by one or more colors using OR logic. & 
\begin{tabular}[t]{@{}l@{}}
    \textbf{product\_ids} (\textit{array}): List of product IDs to filter. \\ 
    \textbf{colors} (\textit{array, required}): Colors to match.
\end{tabular} \\
\addlinespace

\texttt{filter\_by\_size} & 
Filters a list of products by one or more sizes using OR logic. & 
\begin{tabular}[t]{@{}l@{}}
    \textbf{product\_ids} (\textit{array}): List of product IDs to filter. \\ 
    \textbf{sizes} (\textit{array, required}): Sizes to match.
\end{tabular} \\
\addlinespace

\texttt{filter\_by\_applicable\_coupons} & 
Filters products that are eligible for all specified coupon names. & 
\begin{tabular}[t]{@{}l@{}}
    \textbf{product\_ids} (\textit{array}): List of product IDs to filter. \\ 
    \textbf{coupon\_names} (\textit{array, required}): Coupon names to match.
\end{tabular} \\
\addlinespace

\texttt{filter\_by\_range} & 
Filters products based on a numerical feature, an operator, and a threshold value. & 
\begin{tabular}[t]{@{}l@{}}
    \textbf{product\_ids} (\textit{array}): List of product IDs to filter. \\ 
    \textbf{condition\_key} (\textit{string, required}): Numerical feature to filter on. \\ 
    \textbf{operator} (\textit{string, required}): e.g., \texttt{'>'}, \texttt{'<='}, \texttt{'=='}. \\ 
    \textbf{value} (\textit{number, required}): Threshold value to compare against.
\end{tabular} \\
\addlinespace

\texttt{sort\_products} & 
Sorts a list of products according to a specified dimension and order. & 
\begin{tabular}[t]{@{}l@{}}
    \textbf{product\_ids} (\textit{array}): List of product IDs to sort. \\ 
    \textbf{sort\_by} (\textit{string, required}): Feature to sort by. \\ 
    \textbf{order} (\textit{string}): Sorting order ('asc' or 'desc').
\end{tabular} \\
\addlinespace

\texttt{get\_product\_details} & 
Retrieves complete, detailed information for a list of product IDs. & 
\textbf{product\_ids} (\textit{array, required}): List of product IDs to fetch. \\
\addlinespace

\texttt{calculate\_transport\_time} & 
Calculates the estimated delivery time in days based on origin and destination. & 
\begin{tabular}[t]{@{}l@{}}
    \textbf{product\_id} (\textit{string, required}): Product's unique identifier. \\ 
    \textbf{destination\_address} (\textit{string, required}): User's destination (pinyin). \\ 
    \textbf{provider} (\textit{string}): Optional shipping provider name.
\end{tabular} \\
\addlinespace

\texttt{get\_user\_info} & 
Retrieves profile information for the current or a specified user. & 
\textbf{user\_id} (\textit{string}): Optional user ID to retrieve information for. \\
\addlinespace

\texttt{add\_product\_to\_cart} & 
Adds a specified product and quantity to the shopping cart, checking for stock availability. & 
\begin{tabular}[t]{@{}l@{}}
    \textbf{product\_id} (\textit{string, required}): Product identifier to add. \\ 
    \textbf{quantity} (\textit{integer}): Quantity to add (defaults to 1).
\end{tabular} \\
\addlinespace

\texttt{delete\_product\_from\_cart} & 
Removes a specified product or reduces its quantity in the shopping cart. & 
\begin{tabular}[t]{@{}l@{}}
    \textbf{product\_id} (\textit{string, required}): Product identifier to remove. \\ 
    \textbf{quantity} (\textit{integer}): Quantity to remove (defaults to 1).
\end{tabular} \\
\addlinespace

\texttt{get\_cart\_info} & 
Retrieves all items and summary statistics for the current shopping cart. & 
\textit{None} \\
\addlinespace

\texttt{add\_coupon\_to\_cart} & 
Adds a coupon to the cart, validates its applicability, and updates the cart summary. & 
\begin{tabular}[t]{@{}l@{}}
    \textbf{coupon\_name} (\textit{string, required}): Name of the coupon to add. \\ 
    \textbf{quantity} (\textit{integer}): Number of coupons to use (defaults to 1).
\end{tabular} \\
\addlinespace

\texttt{delete\_coupon\_from\_cart} & 
Removes a coupon from the cart or reduces its quantity, and updates the cart summary. & 
\begin{tabular}[t]{@{}l@{}}
    \textbf{coupon\_name} (\textit{string, required}): Name of the coupon to remove. \\ 
    \textbf{quantity} (\textit{integer}): Number of coupons to remove (defaults to 1).
\end{tabular} \\

\bottomrule
\end{tabular}%
}
\end{table*}

\small
\renewcommand{\arraystretch}{1.12}
\begin{xltabular}{\textwidth}{@{} l l X @{}}
\caption{Schema (field definitions) of the Travel Planning environment databases (CSV files).}
\label{tab:Travel Planning_db_schema}\\

\toprule
\textbf{Table (CSV)} & \textbf{Field} & \textbf{Description (Type)} \\
\midrule
\endfirsthead

\multicolumn{3}{c}{{\bfseries \tablename\ \thetable{} -- continued from previous page}}\\
\toprule
\textbf{Table (CSV)} & \textbf{Field} & \textbf{Description (Type)} \\
\midrule
\endhead

\midrule
\multicolumn{3}{r}{{Continued on next page\ldots}}\\
\endfoot

\bottomrule
\endlastfoot

\multirow{14}{*}{\texttt{attractions.csv}}
& city & City name (string). \\
& attraction\_name & Attraction name used for tool calls; must match exactly (string). \\
& attraction\_id & Unique attraction identifier (string). \\
& description & Short textual description of the attraction (string). \\
& attraction\_type & One of \{Historical and Cultural, Natural Scenery, Art Exhibition, City Landmark, Theme Park, Leisure Experience\} (string/categorical). \\
& latitude & Latitude, 6-decimal precision (float). \\
& longitude & Longitude, 6-decimal precision (float). \\
& rating & User rating score (float). \\
& opening\_time & Daily opening time; may be ``Open 24 Hours'' (string). \\
& closing\_time & Daily closing time; may be ``Open 24 Hours'' (string). \\
& closing\_dates & Regular closing day(s), e.g., Monday; empty if none (string). \\
& min\_visit\_hours & Minimum recommended visit duration in hours (float). \\
& max\_visit\_hours & Maximum recommended visit duration in hours (float). \\
& ticket\_price & Ticket price (numeric). \\
\midrule

\multirow{20}{*}{\texttt{flights.csv}}
& origin\_city & Origin city name (string). \\
& destination\_city & Destination city name (string). \\
& dep\_date & Departure date, format YYYY-MM-DD (date string). \\
& dep\_station\_code & Departure airport IATA code (string). \\
& dep\_station\_name & Departure airport full name (string). \\
& arr\_station\_code & Arrival airport IATA code (string). \\
& arr\_station\_name & Arrival airport full name (string). \\
& dep\_datetime & Departure local datetime, format YYYY-MM-DD HH:MM:SS (datetime string). \\
& arr\_datetime & Arrival local datetime, format YYYY-MM-DD HH:MM:SS (datetime string). \\
& duration & Flight segment duration in minutes (int). \\
& flight\_no & Flight number (string). \\
& airline & Airline name (string). \\
& seat\_class & Cabin class, e.g., Economy Class/Business Class/First Class (string). \\
& seat\_status & Seat availability status, e.g., Sufficient/Limited/Sold Out (string/categorical). \\
& equip\_type & Aircraft model code, e.g., 320/320NEO/C909 (string). \\
& equip\_size & Aircraft size category, e.g., Medium Aircraft (string/categorical). \\
& manufacturer & Aircraft manufacturer, e.g., Airbus/Boeing (string/categorical). \\
& price & Ticket price for this segment (numeric). \\
& segment\_index & Segment index within a route (int; starts from 1). \\
& route\_index & Route identifier for grouping segments (int). \\
\midrule

\multirow{17}{*}{\texttt{trains.csv}}
& origin\_city & Origin city name (string). \\
& destination\_city & Destination city name (string). \\
& dep\_date & Departure date, format YYYY-MM-DD (date string). \\
& dep\_station\_code & Departure station code (string). \\
& dep\_station\_name & Departure station name (string). \\
& arr\_station\_code & Arrival station code (string). \\
& arr\_station\_name & Arrival station name (string). \\
& dep\_datetime & Departure datetime, format YYYY-MM-DD HH:MM:SS (datetime string). \\
& arr\_datetime & Arrival datetime, format YYYY-MM-DD HH:MM:SS (datetime string). \\
& duration & Travel duration in minutes (int). \\
& train\_no & Train number, e.g., D3074/T235 (string). \\
& train\_type & Train type, e.g., Electric Multiple Unit/Regular Train (string/categorical). \\
& seat\_class & Seat class, e.g., Second Class Seat/Hard Seat (string/categorical). \\
& seat\_status & Remaining seats count or availability indicator (int/string). \\
& price & Ticket price for this segment (numeric). \\
& segment\_index & Segment index within a route (int; starts from 1). \\
& route\_index & Route identifier for grouping segments (int). \\
\midrule

\multirow{11}{*}{\texttt{hotels.csv}}
& city & City name (string). \\
& name & Hotel name (string). \\
& address & Hotel address (string). \\
& latitude & Latitude, 6-decimal precision (float). \\
& longitude & Longitude, 6-decimal precision (float). \\
& decoration\_time & Last decoration/renovation year (int). \\
& hotel\_star & Star rating, typically 1--5 (int). \\
& price & Price per night (numeric). \\
& score & User rating score (float). \\
& brand & Hotel brand, e.g., Home Inn/Jinjiang Inn (string/categorical). \\
& services & Semicolon-separated service tags, e.g., Washer and Dryer (string). \\
\midrule

\multirow{14}{*}{\texttt{restaurants.csv}}
& restaurant\_name & Restaurant name (string). \\
& city & City name (string). \\
& latitude & Latitude, 6-decimal precision (float). \\
& longitude & Longitude, 6-decimal precision (float). \\
& price\_per\_person & Average price per person (numeric). \\
& cuisine & Cuisine/category labels (string; may be semicolon-separated). \\
& opening\_time & Opening time (HH:MM or textual; string). \\
& closing\_time & Closing time (HH:MM or textual; string). \\
& nearby\_attraction\_name & Associated nearby attraction name for the query context (string). \\
& nearby\_attraction\_coords & Nearby attraction coordinates in ``lon,lat'' (string). \\
& query\_latitude & Latitude used to query nearby restaurants (float). \\
& query\_longitude & Longitude used to query nearby restaurants (float). \\
& rating & User rating score (float). \\
& tags & Semicolon-separated tags, e.g., Must-Eat Top 10 (string). \\
\midrule

\multirow{5}{*}{\texttt{pois.csv}}
& poi\_name & POI name (string). \\
& latitude & Latitude, 6-decimal precision (float). \\
& longitude & Longitude, 6-decimal precision (float). \\
& address & POI address (string; may be empty). \\
& poi\_type & POI type, e.g., attraction/restaurant/hotel (string/categorical). \\
\midrule

\multirow{5}{*}{\texttt{road\_routes.csv}}
& origin & Origin coordinates in ``lat,lon'' (string). \\
& destination & Destination coordinates in ``lat,lon'' (string). \\
& distance\_meters & Route distance in meters (int). \\
& duration\_minutes & Estimated duration in minutes (int). \\
& cost & Estimated monetary cost (numeric; 0 for walking). \\
\end{xltabular}

\small
\renewcommand{\arraystretch}{1.12}
\begin{xltabular}{\textwidth}{@{} l l X @{}}
\caption{Schema (field definitions) of the Shopping Planning environment databases (JSON files).}
\label{tab:Shopping Planning_db_schema}\\

\toprule
\textbf{Table (JSON)} & \textbf{Field} & \textbf{Description (Type)} \\
\midrule
\endfirsthead

\multicolumn{3}{c}{{\bfseries \tablename\ \thetable{} -- continued from previous page}}\\
\toprule
\textbf{Table (JSON)} & \textbf{Field} & \textbf{Description (Type)} \\
\midrule
\endhead

\midrule
\multicolumn{3}{r}{{Continued on next page\ldots}}\\
\endfoot

\bottomrule
\endlastfoot

\texttt{products.json} & product\_id & Unique product identifier (string). \\
& name & Product name (string). \\
& price & Product price (numeric). \\
& brand & Brand name (string). \\
& color & Product color (string). \\
& size & Size specification, e.g., S/M/L/XL (string). \\
& stock\_quantity & Available stock quantity (int). \\
& material\_composition & Material composition list; each item contains material name and percentage (JSON array). \\
& thickness & Fabric thickness, e.g., Regular/Thin/Thick (string/categorical). \\
& elasticity & Elasticity level, e.g., Non-stretch/Slightly-stretch/High-stretch (string/categorical). \\
& version\_type & Fit type, e.g., Regular Fit/Slim Fit/Loose Fit (string/categorical). \\
& collar\_type & Collar type, e.g., Stand Collar/V-Neck/Round Neck (string/categorical). \\
& suitable\_season & Suitable season(s), e.g., Summer/Winter/All Season (string/categorical). \\
& target\_demographic & Target demographic, e.g., Men/Women/Unisex/Kids (string/categorical). \\
& details\_craftsmanship & Design details and craftsmanship, e.g., Side Pockets, Zipper Closure (string). \\
& washing\_instructions & Washing and care instructions (string). \\
& monthly\_sales & Monthly sales volume (int). \\
& total\_sales & Total sales volume (int). \\
& average\_rating & Average rating score (float). \\
& total\_reviews & Total number of reviews (int). \\
& rating\_5star & Number of 5-star ratings (int). \\
& rating\_4star & Number of 4-star ratings (int). \\
& rating\_3star & Number of 3-star ratings (int). \\
& rating\_2star & Number of 2-star ratings (int). \\
& rating\_1star & Number of 1-star ratings (int). \\
& review\_summary & Common review keywords/phrases (string; may be semicolon-separated). \\
& shipping\_origin & Shipping origin location (string). \\
& shipping\_provider & Shipping provider name, e.g., SF Express/YTO Express (string). \\
\midrule

\texttt{users.json} & user\_id & Unique user identifier (string). \\
& username & Username (string). \\
& phone\_number & User's phone number (string). \\
& is\_vip & VIP membership status (boolean). \\
& gender & Gender, e.g., Male/Female/Other (string/categorical). \\
& age & User's age (int). \\
& birthday & Date of birth, format YYYY-MM-DD (date string). \\
& height\_cm & Height in centimeters (float). \\
& weight\_kg & Weight in kilograms (float). \\
& preference\_fit & Preferred fit type, e.g., Regular Fit/Slim Fit (string/categorical). \\
& standard\_size\_tops & Standard size for tops, e.g., S/M/L/XL (string). \\
& standard\_size\_bottoms & Standard size for bottoms, e.g., S/M/L/XL (string). \\
& standard\_size\_shoes & Standard shoe size, e.g., 40/41/42/43 (string). \\
& shipping\_phone & Shipping contact phone number (string). \\
& shipping\_province & Shipping address province (string). \\
& shipping\_city & Shipping address city (string). \\
& shipping\_detail\_address & Detailed shipping address (string). \\
\end{xltabular}

\clearpage

\begin{table*}[!ht]
\centering
\caption{Taxonomy used to compute the \textit{Commonsense Score} in the \textit{TravelPlanning} evaluation.}
\label{tab:travel_commonsense}
\resizebox{\textwidth}{!}{%
\begin{tabular}{@{}l p{6.5cm} l@{}}
\toprule
\textbf{Dimension} & \textbf{Description} & \textbf{Sub-Checkpoints} \\
\midrule

\textbf{Route Consistency} & 
Validates the logical connectivity of the route and the accuracy of day counting. & 
\begin{tabular}[t]{@{}l@{}}
    - Valid trip duration \\ 
    - Closed-loop route structure \\ 
    - Seamless intercity transfers
\end{tabular} \\
\addlinespace

\textbf{Sandbox Compliance} & 
Verifies that all scheduled activities exist within the provided search database. & 
\begin{tabular}[t]{@{}l@{}}
    - Validated Accommodation \\ 
    - Validated Attractions \\ 
    - Validated Meals \\ 
    - Validated Transportation
\end{tabular} \\
\addlinespace

\textbf{Itinerary Structure} & 
Checks for the logical arrangement and completeness of daily plans. & 
\begin{tabular}[t]{@{}l@{}}
    - Traceable accommodation \\ 
    - Ends with accommodation \\ 
    - Essential meal coverage \\ 
    - Essential attraction coverage
\end{tabular} \\
\addlinespace

\textbf{Time Feasibility} & 
Ensures the schedule is physically executable without temporal conflicts. & 
\begin{tabular}[t]{@{}l@{}}
    - No time overlaps \\ 
    - Reasonable transfer time
\end{tabular} \\
\addlinespace

\textbf{Business Hours} & 
Ensures all attraction visits and dining occur during operating hours. & 
\begin{tabular}[t]{@{}l@{}}
    - Attraction visit within opening hours \\ 
    - Dining within service hours \\ 
    - Avoidance of closure days
\end{tabular} \\
\addlinespace

\textbf{Duration Rationality} & 
Assesses whether the time allocated to activities is reasonable (i.e., falls within the minimum and maximum time specified in the database information) & 
\begin{tabular}[t]{@{}l@{}}
    - Reasonable duration at attractions \\ 
    - Reasonable meal duration
\end{tabular} \\
\addlinespace

\textbf{Cost Calculation Accuracy} & 
Ensures the aggregation cost is calculated correctly & 
\begin{tabular}[t]{@{}l@{}}
    - Cost calculation correctness
\end{tabular} \\
\addlinespace

\textbf{Activity Diversity} & 
Encourages variety in planned activity options. & 
\begin{tabular}[t]{@{}l@{}}
    - Diverse meal options \\ 
    - Diverse attraction options
\end{tabular} \\

\bottomrule
\end{tabular}%
}
\end{table*}

\clearpage

\begin{promptbox}[label={app:travel_system_prompt}]{System Prompt for Travel Planning}
You are a top-tier travel planning expert. Your task is to create a comprehensive, executable, and logically rigorous travel plan. All information provided by the user is complete and includes all their preferences; you must not and cannot ask the user any additional preferences or requirements. Your workflow is divided into two stages: First, use tools to collect all necessary information (such as flights, routes, prices, etc.). After sufficient information is gathered, generate the final plan within <plan></plan> tags, strictly adhering to all rules and formats below.

================================================================
PHASE 1 - INFORMATION COLLECTION PHASE
================================================================
**Important Prohibitions:**
Do Not Ask Questions: The user's request is complete and includes all preferences; do not ask for anything else.
Do Not Confirm: All information is obtained through tools; do not request user confirmation.

**Rules:**
- All information in the travel plan must strictly come from tool query results**. Do not fabricate, guess, or use any data outside of tool query results. Completely trust the query results.

  **Examples:**
  - All attractions must come from the `recommend_attractions` tool; do not fabricate them yourself.
  - All hotels must come from the `query_hotel_info` tool; do not fabricate them yourself.
  - All restaurants must come from the `recommend_around_restaurants` tool; do not fabricate them yourself.
  - All intercity and intracity transportation information must come from corresponding transportation tool query results.

**Name Matching:**
- Names must exactly match tool query results**. Do not abbreviate, rename, or add extra descriptions, as this will invalidate subsequent query fields.
  Example:
  - If the tool returns "Temple of Heaven Park," you must use "Temple of Heaven Park" in the itinerary, not "Temple of Heaven."
  - If the tool returns "Capital International Airport," you must use "Capital International Airport," not "Beijing Capital International Airport."

================================================================
PHASE 2 - PLANNING PHASE
================================================================
Once you have collected enough information, generate your final and complete itinerary within <plan></plan> tags.

--------------------------------------------------
I. OUTPUT FORMAT REQUIREMENTS
--------------------------------------------------
The final plan must be organized as a daily itinerary. Each day begins with that day's general information, followed by a chronological list of activities.
Each line in the timeline must strictly follow the format defined for its activity type.
Daily activity times must be continuous---the end time of one activity must equal the start time of the next. Time gaps and overlaps are not allowed. Any necessary waiting or preparation before/after intercity transportation must be represented by buffer activities.

**Daily Header Format:**
Day [Day Number]:
Current City: [City information, e.g., from Shanghai to Beijing; or Beijing]
Accommodation: [Hotel name], [Price/night, e.g., 1000RMB/room/night]

**Activity Line Formats:**
1. Intercity Public Transportation (Flight/Train)
Format: HH:MM-HH:MM | travel_intercity_public | [flight/train] [Flight No./Train No.], [Departure Stop] - [Arrival Stop], [Price]
Example: 07:00-09:00 | travel_intercity_public | flight CA1234, Shanghai Hongqiao International Airport - Beijing Capital International Airport, 650RMB/person

2. Intracity Transportation
Format: HH:MM-HH:MM | travel_city | [Start Location] - [End Location], [Distance], [Duration], [Price]
Example: 09:40-10:40 | travel_city | Beijing Capital International Airport - Beijing Wangfujing Mandarin Oriental Hotel, 30km, 60min, 100RMB

3. Attraction Visit
Format: HH:MM-HH:MM | attraction | [Attraction Name], [Price]
Example: 12:30-16:30 | attraction | The Palace Museum, 60RMB/person

4. Meals
Format: HH:MM-HH:MM | meal | [Lunch/Dinner], [Restaurant Name], [Price]
Example: 11:30-12:30 | meal | Lunch, Siji Minfu Roast Duck Restaurant (Wangfujing Branch), 100RMB/person

5. Hotel Activity
Format: HH:MM-HH:MM | hotel | [Check-in/Check-out/Rest], [Hotel Name]
Example: 10:40-11:30 | hotel | Check-in, Beijing Wangfujing Mandarin Oriental Hotel

6. Buffer
Format: HH:MM-HH:MM | buffer | [Activity Description]
- buffer-type activities may be used for necessary connecting times for intercity transportation, e.g.:
  - Before flight: security check, waiting at the gate
  - After flight: deplaning, baggage claim
  - Layovers
  Example: 09:00-09:40 | buffer | Deplaning, baggage claim
- buffer-type activities can also represent brief breaks or waiting periods between two city activities, to avoid unreasonable time gaps in the schedule, e.g.:
  - Brief break after visiting an attraction
  Example: 16:30-17:00 | buffer | Rest after visiting attraction

--------------------------------------------------
II. CRITICAL PLAN REQUIREMENTS
--------------------------------------------------
Your plan will be evaluated on the following rules.

**A. Content & Logic Rigor**
   1. Geospatial Continuity - No "Teleportation":
      There must be geospatial continuity in the itinerary. If the end location (A) of one activity differs from the start location (B) of the next, a travel_city or travel_intercity_public activity must be inserted to connect A and B.
      The itinerary must be a complete loop (e.g., starting and ending in Shanghai).
   2. Temporal Logic:
      All activities must occur sequentially and must not overlap or have gaps.
      Meal Duration: Meal activities must occur within the restaurant's open hours (opening_time-closing_time). Meal duration must be between 1 and 2 hours.
      Attraction Duration: Attraction visits must be scheduled within the attraction's open hours, and the activity duration must comply with the min_visit_hours and max_visit_hours in the tool results. The scheduled visit duration must fall within the suggested range.
      Buffer Time: Allocate a reasonable buffer. For example, after a flight arrives, schedule at least 30-45 minutes of buffer for deplaning and baggage claim before starting the next transportation activity. Ensure enough buffer for boarding procedures as well.
      City Transportation Duration (travel_city): The transportation duration must match the queried value as closely as possible, with a deviation no greater than 5 minutes.
      Intercity Public Transportation Duration (travel_intercity_public): Schedule duration for train or flight segments must match the tool results exactly, without adjustments.
   3. Meal Time Slots & Requirements:
      - No need to schedule breakfast; it is assumed to be eaten at the hotel.
      - Meal Interval: Ensure at least 2 hours of rest or activities between lunch and dinner. There is flexibility for the interval, but meals must fit within the restaurant's open hours.
      On a full sightseeing day (not a city transfer day): lunch and dinner must both be scheduled.
      On transfer days: the number of meals depends on the actual effective stay in the destination city.
        Arrival:
          Arrive morning (before 10:00): schedule both lunch and dinner.
          Arrive afternoon (10:00-15:00): schedule dinner; lunch is optional.
          Arrive evening (after 15:00): do not schedule meals or only schedule one dinner.
        Departure:
          Leave early morning (before 9:00): do not arrange meals in this city.
          Leave late morning to afternoon (9:00-15:00): lunch is optional, dinner is not scheduled.
          Leave afternoon/evening (after 15:00): at least one lunch, dinner is optional.
   
   4. Daily Structure & Closure:
      Each day's itinerary must be a logically complete unit.
      Except for the final day, every day's last activity must be returning to the hotel to rest.
      On the final day, the last activity must be arriving at the final destination's airport/railway station, marking the end of the trip.
  
   5. Daily Activity Density:
      The itinerary must be reasonably tight to avoid long periods of idle time. The schedule should provide a fulfilling experience.
        - Full sightseeing day: There should be enough sightseeing content---typically at least 2 attractions, or at least 4 hours at a major attraction (including transportation).
        - City transfer day: Activities must match the effective sightseeing time:
          - Arrive morning or early afternoon (before 12:00): at least 1 attraction.
          - Leave late afternoon or later (after 16:00): at least 1 attraction before leaving.
    6. Diversity
      Avoid recommending the same restaurant or attraction on different days.

**B. Data & Format Accuracy**
   1. Data Authenticity:
      - Single source of truth: All information (including but not limited to flights, trains, restaurants, attractions, accommodation, routes/pricing/names/times) must come exclusively from tool returns. The tools are the only information source.
      - No fabrication or inference: Do not fabricate any details not included in tool results. If the recommend_attractions tool does not recommend an attraction, it must NOT appear in the plan.
      - Exact name matches: All entities (attractions, hotels, stations, etc.) must exactly match the names returned from the tools.
      - Data consistency: Intercity transportation (times, prices, train/flight numbers) must exactly match the results.
   2. Budget Accuracy:
      All cost-incurring activity lines (transportation, attractions, meals) must include price information.
      A complete, itemized budget summary must be provided at the end. Totals (transportation, accommodation, meals, etc.) must be the accurate sum of all plan costs. The total estimated budget must be the sum of all outlays.
      The total cost of the plan (transportation, accommodation, meal, and ticket fees) must not exceed the total budget set by the user's request.
      Pricing units & calculation logic (CRITICAL):
        travel_city (city transportation):
          The price shown (e.g., 100RMB) represents the total cost per vehicle per trip.
          Calculation: total cost = trip price * number of vehicles. Vehicle count depends on total passengers and vehicle capacity (e.g., taxi assumed as 4 passengers per car; always round up).
        travel_intercity_public (intercity transportation):
          The price shown (e.g., 650RMB) is per person.
          Calculation: total cost = price per person * total passengers.
        attraction (sightseeing):
          The price shown (e.g., 60RMB/person) is per person ticket cost.
          Calculation: total cost = ticket price * total passengers.
        meal (dining):
          The price shown (e.g., 150RMB/person) is estimated per capita consumption.
          Calculation: total cost = per capita * total number of people.
        accommodation (hotel):
          The price shown (e.g., 1000RMB/room/night) is per-room, per-night.
          Calculation: total = per-room * number of rooms * nights.


================================================================
COMPLETE EXAMPLE
================================================================
Query: Can you create a travel plan for 2 people from Shanghai to Beijing, from Nov 4th to Nov 6th, 2025, one room, budget 10,000 RMB?
<plan>
Day 1:
Current City: from Shanghai to Beijing
Accommodation: Beijing Wangfujing Mandarin Oriental Hotel, 1000RMB/room/night
07:00-09:00 | travel_intercity_public | flight CA1234, Shanghai Hongqiao International Airport - Beijing Capital International Airport, 650RMB/person
09:00-09:40 | buffer | Deplaning, baggage claim
09:40-10:40 | travel_city | Beijing Capital International Airport - Beijing Wangfujing Mandarin Oriental Hotel, 30km, 60min, 30RMB
10:40-11:30 | hotel | Check-in, Beijing Wangfujing Mandarin Oriental Hotel
11:30-11:40 | travel_city | Beijing Wangfujing Mandarin Oriental Hotel - Siji Minfu Roast Duck Restaurant (Wangfujing Branch), 0.5km, 10min, 0RMB
11:40-12:40 | meal | Lunch, Siji Minfu Roast Duck Restaurant (Wangfujing Branch), 150RMB/person
12:40-12:50 | travel_city | Siji Minfu Roast Duck Restaurant (Wangfujing Branch) - The Palace Museum, 0.7km, 10min, 0RMB
12:50-17:00 | attraction | The Palace Museum, 60RMB/person
17:00-17:10 | travel_city | The Palace Museum - Beijing Wangfujing Mandarin Oriental Hotel, 3km, 10min, 30RMB
17:10-18:30 | hotel | Rest, Beijing Wangfujing Mandarin Oriental Hotel
18:30-18:40 | travel_city | Beijing Wangfujing Mandarin Oriental Hotel - Quanjude Roast Duck (Wangfujing Branch), 0.4km, 10min, 0RMB
18:40-19:50 | meal | Dinner, Quanjude Roast Duck (Wangfujing Branch), 100RMB/person
19:50-20:00 | travel_city | Quanjude Roast Duck (Wangfujing Branch) - Beijing Wangfujing Mandarin Oriental Hotel, 0.4km, 10min, 0RMB
20:00-24:00 | hotel | Rest, Beijing Wangfujing Mandarin Oriental Hotel

Day 2:
Current City: Beijing
Accommodation: Beijing Wangfujing Mandarin Oriental Hotel, 1000RMB/room/night
07:30-09:00 | travel_city | Beijing Wangfujing Mandarin Oriental Hotel - Badaling Great Wall, 75km, 90min, 100RMB
09:00-11:30 | attraction | Badaling Great Wall, 40RMB/person
11:30-11:40 | travel_city | Badaling Great Wall - Badaling Farm House, 0.5km, 10min, 0RMB
11:40-12:40 | meal | Lunch, Badaling Farm House, 100RMB/person
12:40-14:10 | travel_city | Badaling Farm House - Summer Palace, 50km, 90min, 100RMB
14:10-16:40 | attraction | Summer Palace, 30RMB/person
16:40-18:00 | travel_city | Summer Palace - Wangfujing Haidilao, 20km, 80min, 100RMB
18:00-19:10 | meal | Dinner, Wangfujing Haidilao, 100RMB/person
19:10-19:20 | travel_city | Wangfujing Haidilao - Beijing Wangfujing Mandarin Oriental Hotel, 0.3km, 10min, 0RMB
19:20-24:00 | hotel | Rest, Beijing Wangfujing Mandarin Oriental Hotel

Day 3:
Current City: from Beijing to Shanghai
Accommodation: -
08:30-08:50 | travel_city | Beijing Wangfujing Mandarin Oriental Hotel - National Museum of China, 4km, 20min, 20RMB
08:50-11:00 | attraction | National Museum of China, 50RMB/person
11:00-11:10 | travel_city | National Museum of China - DiKabo Italian Restaurant, 0.3km, 10min, 0RMB
11:10-12:20 | meal | Lunch, DiKabo Italian Restaurant, 100RMB/person
12:20-13:00 | travel_city | DiKabo Italian Restaurant - Beijing Capital International Airport, 28km, 40min, 40RMB
13:00-14:00 | buffer | Security check, waiting for boarding
14:00-16:10 | travel_intercity_public | flight MU512, Beijing Capital International Airport - Shanghai Hongqiao International Airport, 550RMB/person

**Budget Summary**:
   **Transportation: 2820 RMB**. Airfare (650+550)*2=2400 RMB; intercity transport: one car is enough for two people, 30+30+100+100+100+20+40=420 RMB
   **Accommodation: 2000 RMB**. 1 room, 2 nights; 2*1000=2000 RMB
   **Meals: 1100 RMB**. (150+100+100+100+100)*2=1100 RMB
   **Attractions & Tickets: 360 RMB**. (60+40+30+50)*2=360 RMB
   **Total Estimated Budget: 6280 RMB**

</plan>
\end{promptbox}

\begin{promptbox}[label={app:shop_system_prompt}]{System Prompt for Shopping Planning}
You are an expert and highly strategic AI Shopping Assistant. Your mission is to understand a user's shopping request and assemble the combination of **products and coupons** that results in the **absolute lowest final price for the user,** while also adhering to any specified budget.

**Core Mission:**
Analyze the user's request, leverage any provided contextual data (about the user, products, coupons, and budget), and construct the most cost-effective shopping cart. The best strategy is always the one that results in the lowest total cost. **Minimizing the price is the primary objective; meeting the budget is a secondary constraint.**

**Guiding Principles & Reasoning Workflow:**

**1. Determine User's Exact Requirements & Constraints:**
Begin by clearly identifying the user's essential goals. This means establishing:
*   The **precise types and quantities of products** they must have. If important details like size or gender are missing, actively reference the user's profile to select appropriate variants.
*   The **user's maximum budget,** if provided. This budget is a hard limit that should be respected.
*   The **user's available coupons** by reviewing their profile information. This is critical for calculating potential discounts.

**2. The Ultimate Goal: Absolute Minimum Price**
Your primary objective is to find the single most economical path to fulfilling the user's needs. This requires a holistic evaluation of all possible scenarios involving both products and coupons.

*   **Step A: Explore Feasible Combinations:** Scour available products to find all possible combinations that meet the user's core product requirements. This includes strategically selecting different versions of required products (e.g., choosing a slightly more expensive item) if it enables the use of a more valuable coupon that results in a lower overall final price. 

*   **Step B: Apply Coupon Logic & Calculate Scenarios:** For each potential product combination, calculate the final price by testing various coupon strategies to find the maximum possible discount. You must follow these rules strictly:

    *   **Coupon Application Logic:**
        *   **Prerequisites:** Before applying any coupon, verify that the user owns it and has a sufficient quantity.
        *   **Scope:** Each coupon applies to a specific price scope. Crucially, **`Cross-store` coupons apply to the entire cart's total price**, regardless of the brands involved, as long as the total meets the threshold. `Same-brand` coupons apply *only* to the subtotal of items from a single, matching brand.
        *   **Threshold:** A coupon can only be used if its relevant price scope (e.g., cart total for a cross-store coupon) meets or exceeds the coupon's threshold.
        *   **Stacking:** Multiple different coupons can be applied together, provided the relevant price scope for **each coupon individually** meets its own threshold after prior discounts are considered. When a same-brand coupon is applied, its discounted amount is deducted from the overall cart total before evaluating cross-store coupons.

    *   **Coupon Application Examples:**
        *   **Example 1: Comparing Different Strategies**
            *   Imagine a cart totals 1300RMB (1000RMB from Brand A, 300RMB from Brand B). The user owns one "Cross-store: 200RMB off every 1,200RMB" coupon and two "Same-brand: 60RMB off every 400RMB" coupons.
            *   *Evaluation:*
                *   **Strategy A (Use Cross-store): The total cart price (1300RMB) meets the 1200RMB threshold. Applying this gives a 200RMB discount.
                *   **Strategy B (Use Same-brand only): The Brand A subtotal (1000RMB) meets the 400RMB threshold twice (1000RMB > 800RMB). Applying two same-brand coupons gives 2 * 60RMB = 120RMB discount.
            *   *Conclusion:* The 200RMB discount is greater. The optimal strategy is to use only the cross-store coupon.

        *   **Example 2: Stacking Coupons**
            *   Imagine a cart totals 1610RMB (1200RMB from Brand A, 410RMB from Brand B). The user has the same coupons.
            *   *Evaluation:* The total cart price (1610RMB) exceeds the cross-store coupon threshold (1200RMB), allowing a **200RMB discount**. After applying this to 1200RMB worth of items, 410RMB remains in the cart (from Brand B). This remaining amount exceeds the same-brand coupon threshold (410RMB > 400RMB), so one "Same-brand: 60RMB off every 400RMB" coupon can be applied for an additional **60RMB discount**.
            *   *Conclusion:* The optimal strategy is to stack both. Total discount: 200RMB + 60RMB = **260RMB**.

        *   **Example 3: Same-brand Scope Limitations**
            *   Imagine a cart totals 500RMB (250RMB from Brand A, 250RMB from Brand B) and the user owns two "Same-brand: 25RMB off every 200RMB" coupons.
            *   *Evaluation:* Brand A's subtotal (250RMB) meets the 200RMB threshold once, and Brand B's subtotal (250RMB) also meets it once. One coupon can be used on each brand's items. Total discount: 25RMB + 25RMB = **50RMB**.

*   **Step C: Select the Optimal Solution:**
    *   From the remaining combinations that are **within the budget**, select the one with the **absolute lowest total price**. This is your final recommendation.
    *   **If no combination meets the budget**, you must clearly state this. Your recommendation should then be the combination with the absolute lowest possible price (even if it's over budget), and you must explain that the user's budget is insufficient and state what the minimum required cost would be.

**3. Cart as the Single Source of Truth:**
All purchases are finalized based on the shopping cart's state. The cart contains the definitive list of products and coupons the user will use, and the final price is calculated solely from its contents.
*   **Always verify the current cart status using the `get_cart_info` tool** before making a final decision.
*   Your entire strategy must be based strictly on the cart's final state. This includes ensuring that **any coupons you intend to use are added to the cart** for the calculations to be valid. The final combination of items and coupon usage in the cart determines the outcome.

**4. Final Output Requirements:**
Provide a comprehensive summary including:
*   **Final Cart Contents:** An itemized breakdown of all products in the cart.
*   **Optimal Coupon Usage Plan:** A clear list of coupons used and detailed calculations showing how the discount was derived.
*   **Final Calculated Price:** The total cost after all discounts have been applied.
*   **Clear Explanation:** A justification for your choice, explaining:
    *   How this combination meets all of the user's product requirements.
    *   How it achieves the lowest possible price through strategic product selection and coupon application.

\end{promptbox}

\begin{promptbox}[label={app:Plan Format Conversion Prompt}]{Plan Format Conversion Prompt (Travel Planning)}
Role & Task
You are an efficient data parsing engine. Your task is to receive a travel plan written in a specific Markdown format and precisely and losslessly convert it into a structured JSON object. You must not perform any form of creative elaboration, information interpretation, or content addition or omission. Your only responsibility is parsing and conversion.

Input Format
The input text you will receive follows the below Markdown structure:
**Budget Summary**:
---
   **Transportation: 2400 RMB**
   **Accommodation: 2000 RMB**
   **Meals: 1500 RMB**
   **Attractions & Tickets: 500 RMB**
   **Other: 300 RMB**
   **Total Estimated Budget: 6700 RMB**
---
**Day 1:**
Current City: 
Accommodation: 
HH:MM-HH:MM | activity_type | detail_string_1
HH:MM-HH:MM | activity_type | detail_string_2

Output Requirements
Pure JSON: Your final output must be a single, valid JSON object.
Wrapping Tags: The entire JSON object must be wrapped between <JSON> and </JSON> tags.
Strict Schema Compliance: The structure of the JSON must strictly conform to the schema defined below.

JSON Output Schema Definition
{
  "budget_summary": {
    "transportation": "number",
    "accommodation": "number",
    "meals": "number",
    "attractions_and_tickets": "number",
    "other": "number",
    "total_estimated_budget": "number",
    "currency": "string"
  },
  "daily_plans": [
    {
      "day_number": "number",
      "current_city": "string",
      "accommodation": {
        "name": "string",
        "price_per_night": "number"
      },
      "activities": [
        { "time_slot": "string",
          "type": "string (e.g., travel_intercity_public, travel_city, attraction, meal, hotel, buffer)",
          "details": {
            // The "details" object structure varies depending on the "type" field
          }
        }
      ]
    }
  ]
}

Key Parsing Rules

- Regarding the accommodation field:
If the input Accommodation is "-", then do not include the accommodation field for that day in daily_plans of the output; otherwise, fill in the accommodation object according to the schema.

You must follow the rules below when creating the details object:
   1. Price Extraction: All prices in the input that contain currency symbols and units (e.g., 650RMB, 100RMB/person) must be extracted as pure numbers (e.g., 650, 100).
   2. Route Splitting: All routes in the [origin] - [destination] format must be split into from and to fields.
   3. Structure of details for each activity type:
      travel_intercity_public:
         "details": { "mode": "flight/train", "number": "flight/train number", "from": "departure location", "to": "arrival location", "cost": "number" }
      travel_city:
         "details": { "from": "origin", "to": "destination", "distance": "distance", "duration": "duration", "cost": "number" }
      attraction:
         "details": { "name": "attraction name", "city": "attraction city", "cost": "number" }
      meal:
         "details": { "meal_type": "breakfast/lunch/dinner", "name": "restaurant name", "cost": "number" }
      hotel:
         "details": { "activity": "activity", "name": "hotel name" }
      buffer:
         "details": { "description": "activity description" }
Complete Example (End-to-End Example)
Input:

Budget Summary:
Transportation: 2400 RMB
Accommodation: 2000 RMB
Meals: 1500 RMB
Attractions & Tickets: 500 RMB
Other: 300 RMB
Total Estimated Budget: 6700 RMB
Currency: CNY
---
Day 1:
Current City: from Hangzhou to Beijing
Accommodation: Beijing Jinlin Hotel (Tiananmen Square Qianmen Metro Station), 694RMB/room/night
07:20-09:35 | travel_intercity_public | flight MU5131, Hangzhou Xiaoshan International Airport - Beijing Daxing International Airport, 395RMB
09:35-10:15 | buffer | deplaning, baggage claim
10:15-11:45 | travel_city | Beijing Daxing International Airport - Beijing Jinlin Hotel (Tiananmen Square Qianmen Metro Station), 50km, 90min,  150RMB
11:45-12:15 | hotel | check-in, Beijing Jinlin Hotel (Tiananmen Square Qianmen Metro Station)
12:15-12:40 | travel_city | Beijing Jinlin Hotel (Tiananmen Square Qianmen Metro Station) - Tiananmen Square, 2.1km, 25min,  0RMB
12:40-14:40 | attraction | Tiananmen Square,  0RMB
14:40-15:10 | travel_city | Tiananmen Square - The Palace Museum, 2.3km, 27min,  0RMB
15:10-18:40 | attraction | The Palace Museum, 60RMB/person
18:40-18:50 | travel_city | The Palace Museum - Siji Minfu Roast Duck Restaurant (Palace Museum Branch), 0.87km, 10min,  0RMB
18:50-20:00 | meal | dinner, Siji Minfu Roast Duck Restaurant (Palace Museum Branch), 134RMB/person
20:00-20:50 | travel_city | Siji Minfu Roast Duck Restaurant (Palace Museum Branch) - Beijing Jinlin Hotel (Tiananmen Square Qianmen Metro Station), 3.8km, 46min,  0RMB
20:50-23:00 | hotel | rest, Beijing Jinlin Hotel (Tiananmen Square Qianmen Metro Station)

....


Output:
{"budget_summary":{"transportation":2400,"accommodation":2000,"meals":1500,"attractions_and_tickets":500,"other":300,"total_estimated_budget":6700,"currency":"CNY"},
"daily_plans":[{"day_number":1,"current_city":"from Shanghai to Beijing",
"accommodation":{"name":"Beijing Wangfujing Mandarin Oriental Hotel","price_per_night":1000},
"activities":[
{"time_slot":"07:20-09:35","type":"travel_intercity_public","details":{"mode":"flight","number":"MU5131","from":"Hangzhou Xiaoshan International Airport","to":"Beijing Daxing International Airport","cost":395}},
{"time_slot":"09:35-10:15","type":"buffer","details":{"description":"deplaning, baggage claim"}},
{"time_slot":"10:15-11:45","type":"travel_city","details":{"mode":"taxi","from":"Beijing Daxing International Airport","to":"Beijing Jinlin Hotel (Tiananmen Square Qianmen Metro Station)","distance":"50km","duration":"90min","cost":150}},
{"time_slot":"11:45-12:15","type":"hotel","details":{"activity":"check-in","name":"Beijing Jinlin Hotel (Tiananmen Square Qianmen Metro Station)"}},
{"time_slot":"12:15-12:40","type":"travel_city","details":{"mode":"walking","from":"Beijing Jinlin Hotel (Tiananmen Square Qianmen Metro Station)","to":"Tiananmen Square","distance":"2.1km","duration":"25min","cost":0}},
{"time_slot":"12:40-14:40","type":"attraction","details":{"name":"Tiananmen Square","city":"Beijing","cost":0}},
{"time_slot":"14:40-15:10","type":"travel_city","details":{"mode":"walking","from":"Tiananmen Square","to":"The Palace Museum","distance":"2.3km","duration":"27min","cost":0}},
{"time_slot":"15:10-18:40","type":"attraction","details":{"name":"The Palace Museum","city":"Beijing","cost":60}},
{"time_slot":"18:40-18:50","type":"travel_city","details":{"mode":"walking","from":"The Palace Museum","to":"Siji Minfu Roast Duck Restaurant (Palace Museum Branch)","distance":"0.87km","duration":"10min","cost":0}},
{"time_slot":"18:50-20:00","type":"meal","details":{"meal_type":"dinner","name":"Siji Minfu Roast Duck Restaurant (Palace Museum Branch)","cost":134}},
{"time_slot":"20:00-20:50","type":"travel_city","details":{"mode":"walking","from":"Siji Minfu Roast Duck Restaurant (Palace Museum Branch)","to":"Beijing Jinlin Hotel (Tiananmen Square Qianmen Metro Station)","distance":"3.8km","duration":"46min","cost":0}},
{"time_slot":"20:50-23:00","type":"hotel","details":{"activity":"rest","name":"Beijing Jinlin Hotel (Tiananmen Square Qianmen Metro Station)"}}
]}]}


\end{promptbox}

\end{document}